\documentclass[longtitle,sort&compress]{elsarticle}

\makeatletter
\def\ps@pprintTitle{%
 \let\@oddhead\@empty
 \let\@evenhead\@empty
 \def\@oddfoot{\begin{minipage}[t]{\textwidth}
                       \footnotesize \textcopyright 2018. This manuscript version is made available under the CC-BY-NC-ND 4.0 license  \\                                                                                                    
                                          \url{http://creativecommons.org/licenses/by-nc-nd/4.0/}\\
                                          For the final article published in Elsevier's \emph{Swarm and Evolutionary Compputation} refer to\\ \url{https://dx.doi.org/10.1016/j.swevo.2018.10.002}
                \end{minipage}                       
                }
 
 \let\@evenfoot\@oddfoot}
\makeatother
  
  \usepackage[utf8]{inputenc}
  \usepackage[hidelinks]{hyperref}
  \usepackage[ngerman,american]{babel}
  \usepackage{graphicx}
  \usepackage{algorithmicx}
  \usepackage{algorithm}
  \usepackage{algpseudocode} 
  \usepackage{bibentry}
  \usepackage{todonotes}
  \usepackage{bm}
  \usepackage{caption}
  \usepackage{amsmath,amssymb,dsfont}
  \usepackage{float}
  \usepackage{placeins}
  \usepackage{color}
  \usepackage{mathtools}
  \usepackage{footnote}
  \makesavenoteenv{tabular}
  \makesavenoteenv{table}

\begin{document}

  \begin{frontmatter}

    \title{Benchmarking Evolutionary Algorithms For {Single Objective} Real-valued Constrained Optimization\\ -- A Critical Review -- }

   \author[fhv]{Michael~Hellwig\corref{cor1}}  
    \ead{Michael.Hellwig@fhv.at}
    \cortext[cor1]{Corresponding author}
    \author[fhv]{Hans-Georg~Beyer}
    \ead{Hans-Georg.Beyer@fhv.at}
    \address[fhv]{Vorarlberg University of Applied Science, Research Centre PPE,\\ Campus V, Hochschulstraße 1, 6850 Dornbirn, Austria.}
    
        \begin{abstract}
        Benchmarking plays an important role in the development of novel search algorithms as well as for the assessment and comparison of contemporary algorithmic ideas. This paper presents common principles that need to be taken into account when considering benchmarking problems for constrained optimization. 
	  Current benchmark environments for testing Evolutionary Algorithms are reviewed in the light of these principles.
	  Along with this line, the reader is provided with an overview of the available problem domains in the field of constrained benchmarking.
	  Hence, the review supports algorithms developers with information about the merits and demerits of the available frameworks.
      \end{abstract}


  \end{frontmatter}

\section{Introduction}

Representing a subclass of derivative-free, nature-inspired methods for optimization,
Evolutionary Algorithms (EA) provide powerful optimization tools for, but not restricted to, black-box or simulation-based optimization problems. That is, for problems where the analytical structure of the optimization problem is unknown by default. EA applications to such problems can be found in the fields of operations research, engineering, or machine learning~\cite{michalewicz1996evolutionary,Oduguwa2005,Zhang2011,collange2010multidisciplinary,mora2015applications}.

Due to the lack of theoretical performance results for optimization tasks of notable complexity, the development and the performance comparison of EA widely rely on benchmarking. First and foremost, benchmarking experiments are established for performance evaluation and algorithm comparison on given problem classes.
Ideally, this is supposed to support the selection of the algorithm best suitable for given real-world applications~\cite{Mersmann2015}.
Yet, benchmarks can also be used to experimentally provide insight into the working principles of an algorithm (although, usually purpose-built experiments have to be conducted in addition) and foster the development of algorithms for specific problem branches. Furthermore, benchmarks may qualify to verify theoretical predictions of the algorithm behavior~\cite{Whitley1996,rardin2001experimental}.

Currently, there are basically two main developing lines for EA benchmarking, the test environments provided in the IEEE Congress on Evolutionary Computation (CEC) competitions and the Comparing Continuous Optimizer (COCO) benchmark suite.

The COCO suite~\cite{hansen2016coco} represents the most elaborated platform for 
benchmarking and comparing {unconstrained} continuous optimizers for numerical (non-linear) optimization. The COCO framework advanced from the Black-box Optimization Benchmarking (BBOB) 2009 benchmark set~\cite{HansenAFR2009,hansen:inria-00362633}.
The platform provides tools to ease the process of quantifying and comparing the performance of
optimization algorithms for single-objective noiseless and noisy problems, and for bi-objective noiseless problems, respectively.
A particular strength of the COCO platform is the large number of algorithm results available for comparison. Up to now, 231 distinct (classical as well as contemporary) algorithms have been tested on the COCO testbeds.

Alternatively, the competitions that are organized on a yearly basis during the CEC aim at the comparison of state-of-the-art stochastic search algorithms. These competitions, among others, include single objective, large-scale, noisy, multi-objective, and constrained optimization problems, respectively.
The CEC competitions provide specific test environments for algorithm assessment and comparison. The test function environments made available in this context turned out very popular for benchmarking Evolutionary Algorithms (EA).

Being commonly recognized as successful optimization strategies in the context of unconstrained optimization,
the application of EA to constrained optimization problems has gained the attention of the research community in recent years. 
Constrained optimization tasks are concerned with searching for the optimal solution of an objective function with respect to 
limitations on the search space parameter vector. 
In many real-world applications, constraints result from physical boundaries on the input data, from considering problem specific trade-offs, 
or from limiting the resources of a problem. Regardless of the sources, the introduction of constraints increases the complexity of an 
optimization task. This is particularly true in the context of black-box and simulation-based optimization.
Refer to~\cite{MEZURAMONTES2011} for a survey on commonly used constraint handling approaches in the context of nature-inspired algorithms.
Taking into account constrained optimization problems, the theoretical background of EA is even less developed. Hence, usage of benchmarks for performance assessment and algorithm development is essential. 

Regarding EA benchmarks for constrained optimization, the CEC competitions on constrained real-parameter optimization~\cite{CEC2006,CEC2010,CEC2017} (organized in 2006, 2010, and 2017) introduced specific constrained test environments. The constrained test functions included in the CEC 2006 benchmark definitions were collected from~\cite{MichalewiczS1996,himmelblau1972applied,floudas1999handbook,Xia96,epperly1995global}. The following competitions refined some benchmark definitions and introduced new problem instances. To this end, the test-case generator developed in~\cite{Michalewicz2000} was called on. The respective paper introduces a method to generate test problems with varying features, e.g. with respect to the problem size, the size of the feasible region, or the number and the type of the constraints. Benchmark problems created by the test-case generator are included into the CEC2010 and CEC2017 competition on constrained real-parameter optimization. Until today, the CEC function sets are most frequently used for benchmarking contemporary EA in the context of constrained optimization. Refer to Sec.~\ref{cec} for the review of the CEC benchmark sets.

Only recently, the development of a COCO branch for constrained black-box optimization benchmark ({BBOB-constrained}) problems is near completion~\cite{CocoCode}\footnote{The code related to the BBOB-constrained suite under development is available in the \texttt{development} branch on the project website \url{http://github.com/numbbo/coco/development}.}.
Although the BBOB-constrained suite is still under development, a review of the corresponding benchmark principles is provided in Sec.~\ref{coco}.
Including the unfinished BBOB-constrained suite into the considerations is reasonable for the following reason. 
Representing the most sophisticated framework for unconstrained benchmarking problems, the COCO related benchmarking principles clearly add value to the present discussion. Being near completion, substantial changes to the BBOB-constrained suite are not expected anymore.

An overview of additional problem collections is available at~\cite{NeumaierGlobalOpt}.
Each of these test problem sets is useful for demonstrations of the applicability of interesting algorithmic ideas. However, the presented problems are mainly related to the field of mathematical programming. 
They are provided in different mathematical modeling systems like GAMS\footnote{GAMS -- General Algebraic Modeling System, \url{https://gams.com}} or AMPL\footnote {AMPL -- A Mathematical Programming Language, \url{https://ampl.com}}, but also Fortran, C, or MATLAB implementations exist.
While providing a large number of test problems, the collections commonly leave the initial problem collection as well as the post-processing of algorithm results to the user. Concentrating on deterministic solvers, there do not exist guidelines for experimental design, e.g. in terms of repetitions, or counting function evaluations, respectively.
The absence of a consistent experimental framework often restrains such test function collections from allowing for broad conclusions with respect to algorithm performance. Furthermore, most test problems are designed with a 2fixed search space dimension and a fixed number of constraints which directly impedes their scalability.

Similar concerns apply to many real-world problem applications that are present in the literature~\cite{rardin2001experimental}. They usually come with limited reproducibility and comparability of the results reported, e.g. due to unavailable data or implicit modeling assumptions. Consequently, these studies can rather be thought of as a demonstration of the applicability of a certain algorithm in a particular context than a proof of its superiority. The focus is more on the algorithm output than on the algorithm efficiency~\cite{johnson2002}.

The main goal of the present paper is to provide a critical review of state-of-the-art benchmarking environments that can be used for assessing and comparing Evolutionary Algorithms in the context of constrained single objective real-valued optimization. 
To this end, existing benchmark principles for constrained optimization are collected and complemented with insights from the context of other benchmarking fields and experimentation. Current benchmarking environments are surveyed in the light of these rationales.\footnote{Note that the classification of the great number of solitary test problems~\cite{NeumaierGlobalOpt} is not considered the purpose of this review. Instead, the focus is on the most elaborated constrained benchmarking environments for Evolutionary Algorithms, i.e. the constrained test environments established for the CEC competitions as well as the COCO BBOB-constrained suite.}
This way, the article may raise awareness of recent benchmarking techniques as well as their corresponding strengths and their incapabilities.
By suggesting room for improvements with respect to framework definitions, experimentation principles, and reporting styles, 
the present paper aims at stimulating the debate on benchmarking principles for constrained real-valued optimization.

Such a discussion seems necessary as the field of constrained optimization is spacious and the available benchmarking approaches are comparably scarce. Although some investigations exist~\cite{mezura2004makes}, it is by no means conclusively determined which features are making a constrained optimization problem hard even for a single algorithm subclass. 
Constrained real-valued optimization problems may differ with respect to the following features (and their combinations), including but not necessarily restricted to, \begin{itemize}
      \itemsep-1pt
     \item the number of constraints,
     \item the type of the constraints (refer to Sec.~\ref{problem}),
     \item the analytical structure of objective function and constraints, e.g.
	\begin{itemize}
	  \itemsep-2pt
	  \item the conditioning of the problem
	  \item the modality of the objective function
	  \item the ruggedness of the objective function
	  \item the (non-)linearity of the constraints
	  \item the separability of the objective function and/or constraints
	  \item the number of global optima (inside the feasible region)
	\end{itemize}  
     \item the size of the search space,
     \item the relative size of the feasible region in the search space,
     \item the connectedness of the feasible region,
     \item the orientation of the feasible region within the search space,
     \item the location of the global optimum on the boundary or aside.
    \end{itemize}
As there certainly is no such thing as free lunch~\cite{NFL1997}, and as the EA development for constraint optimization tasks will further rely on the availability of suitable benchmarks, the need for benchmark definitions that take into account consistent subgroups of conceivable problems is beyond dispute.
The test-case generator introduced in~\cite{Michalewicz2000} or test problem collections like~\cite{CUTEst2018}, can be regarded as a meaningful step towards creating well structured problem groups of distinct characteristics. However, these problems and their reported solutions are commonly not scalable with respect to the problem dimensionality.
Further, the issue of proposing a well-defined benchmarking framework as well as providing coherent reporting and ranking principles for meta-heuristic algorithms is not in the scope of these test suites.

The remainder of this paper is organized as follows: Section~\ref{problem} introduces the general real-valued constrained optimization problem~\eqref{cop}, particularly with regard to a classification of the constraint functions commonly used for benchmarking EA on black-box problems. Section~\ref{sec3} presents benchmarking principles appropriate for the comparison of constrained optimization benchmarks. Afterward, the design of the CEC test function sets for constrained optimization and the COCO BBOB-constrained suite are presented in greater detail in Sec.~\ref{cec}, and Sec.~\ref{coco}, respectively.
Both sections particularly expand on the proirly motivated principles.
A discussion of the recent benchmarking principles for EA on constrained optimization problems concludes the paper in Sec.~\ref{discussion}.

\section{Problem formulation}
\label{problem}
  The present paper focuses on continuous optimization problems. That is, both the objective function 
  and the constraint functions are assumed to be real-valued functions.
  The objective function might either be represented as a \emph{reward} or as a \emph{cost function}.
  While the former calls for maximization, a cost function representation needs to be minimized.
  Some collections of benchmark functions may even use both representations, but this paper without loss of generality focuses on minimization problems.

  The constraint functions fall into even more classes. 
  A detailed taxonomy of constraints is provided in~\cite{LedWild2015}. 
  The paper subdivides constraints into nine distinct constraint classes which rely on the categorization according to the following features.
  \begin{description}
  \item[Non-/Quantifiable] A constraint is said to be \emph{quantifiable} if its degree of feasibility and/or constraint violation can be determined. Otherwise, the constraint is denoted \emph{nonquantifiable}. {That is, \emph{nonquantifiable} constraints may only return a boolean expression regarding a constraint's feasibility.}
  \item[Un-/Relaxable] \emph{Unrelaxable} constraints define conditions for the parameter vectors that are required to be satisfied to obtain meaningful outputs from either objective functions or simulations. In contrast, \emph{relaxable} constraints represent desired conditions which do not have to be satisfied at each stage of the optimization process. 
  \item[A priori/Simulation] In case that the feasibility of a constraint can be evaluated directly, it is referred to as \emph{a priori} constraint. A constraint that requires running a simulation to verify its feasibility is denoted a \emph{simulation} constraint.    
  \item[Known/Hidden] While \emph{hidden} constraints are unknown to a solver, \emph{known} constraints are explicitly stated in the problem formulation and thus available to the solver. Notice, \emph{hidden} constraints are distinctive of simulation-based optimization problems. They are \emph{nonquantifiable} and \emph{unrelaxable} by definition. 
  \end{description}
  All combinations of the above categories are reasonable for the definition of problem instances for a constrained benchmark problem.
  However, the benchmark suites considered in this paper are usually dedicated to \emph{known}, \emph{a-priori}, and \emph{quantifiable} constraints.
  Whether some constraints are \emph{relaxable} or \emph{unrelaxable} is depending on the respective benchmark definitions.
  {Particularly, the {(in-)}equality constraint definitions provided in the CEC and COCO constrained benchmarks must be considered Quantifiable/Relaxable/A-priori/Known (or simply QRAK) constraints in this taxonomy. Further, the related box-constraints which are assumed to be satisfied prior to the evaluation of a candidate solution, represent an example of QUAK constraints (Quantifiable/Unrelaxable/A-priori/Known). Refer to~\cite{LedWild2015} for a more detailed explanation.}
  
    Note that, by interpreting all constraints as \emph{unrelaxable}, the problem instances would become considerably harder for EA to satisfy.
    That is, suitable algorithms would have to be equipped with a sophisticated repair technique that allows generating usable candidate solutions in every situation. 
  
  The real-valued constrained optimization problems (COP) considered in this report have the general representation
      \begin{equation}
      \tag{COP}
	\begin{aligned}
	\min \:\: & f(\bm{y})  & &\\
	  s.t. \:\: & g_i(\bm{y}) \leq 0,  & i=1,\dots,l,\: &\\
	      & h_j(\bm{y}) = 0,     & j=1,\dots,k, &\\
	      & \bm{y} \in \mathcal{S} \subseteq \mathds{R}^N. & &
	\end{aligned}
	\label{cop}
      \end{equation}
  In this context, $\bm{y} \in \mathcal{S}$ denotes the $N$-dimensional search space parameter vector. 
  The set $\mathcal{S}$ usually comprises a number of box-constraints specifying reasonable 
  intervals of the parameter vector components, i.e.
  \begin{equation}
  \label{boxset}
    \mathcal{S} \coloneqq \left\{ \bm{y} \in \mathds{R}^N \: | \: \bm{\check{y}} \preceq \bm{y} \preceq \bm{\hat{y}}  \right\},
  \end{equation}
  where the vector $\bm{\check{y}} \in \mathds{R}^N $ consists of the component-wise lower bounds, and $\bm{\hat{y}}$ the vector of
  upper bounds, respectively. Note that, $\preceq$ is understood as the component-wise less than or equal inequality.
  The set $\mathcal{S}$ is also referred to as the box of problem~\eqref{cop}.

  The feasible region of the search space is additionally restricted by $m=l+k$ real-valued constraint functions.
  These constraint functions are separated into $l$ inequality constraints $g_i(\bm{y}), \: i=1,\dots,l$, and $k$
  equality constraints $h_j(\bm{y}), \: j=1,\dots,k$. A vector $\bm{y}\in \mathcal{S}$ that satisfies all constraints is called feasible. 
  The set of all feasible parameter vectors is referred to as 
  \begin{equation}
  \label{feasSet}
    \mathcal{M} \coloneqq \left\{ \bm{y} \in \mathcal{S} \: | \:  g_i(\bm{y}) \leq 0 \:\wedge\:  h_j(\bm{y}) = 0, \: \forall i=1,\dots,l,\: j=1,\dots,k \right\}. 
  \end{equation}
  The global optimum of~\eqref{cop} is denoted by $\bm{y}^* \in \mathcal{M}$.
  Note that the objective function $f(\bm{y})$ subject to some constraints is also referred to as \emph{constrained function}.
  Multiple representations of one specific constrained function that are subject to small variations are denoted as \emph{instances} of that respective constrained function. Such variations involve, for example, the orientation of the feasible region, negligible change in the size of the feasible region or the location of the optimum. Contrary,~\eqref{cop} instances are similar with respect to the objective functions as well as number and analytical type of the constraint functions.\footnote{Aiming at a consistent terminology for the remainder of this article, this denotation of a constrained problem instance does not demand generality.}

  The box-constraints which impose restrictions on the parameter vector components are usually considered unrelaxable.
  On the contrary, inequality and equality constraints are considered relaxable insofar as
  the constrained functions can be evaluated for infeasible parameter vectors and such infeasible candidate solutions may also be employed in the search process.

  The size of the feasible region $\lvert\mathcal{M}\rvert$ relative to the box size $\lvert\mathcal{S}\rvert$ is denoted by
  \begin{equation}
    \label{relsize}
    \rho = \cfrac{\lvert\mathcal{M}\rvert}{\lvert\mathcal{S}\rvert}
  \end{equation}
  The parameter $\rho$ can be estimated by uniformly sampling a sufficiently large number of candidate solutions inside the set $\mathcal{S}$ and by counting the feasible candidate solutions among these, as suggested in~\cite{Koziel1999}.
  
  Considering problem~\eqref{cop}, evolutionary algorithms employ a measure of infeasibility to guide the search process into feasible regions of the search space. 
  The constraint violation $\nu({\bm{y}})$ of a candidate solution $\bm{y}$ is usually specified as
  \begin{equation}
  \begin{split}
    \nu({\bm{y}}) = 0, & \textrm{ if } \bm{y} \in \mathcal{M}, \\ 
    \nu({\bm{y}}) > 0, & \textrm{ if } \bm{y} \notin \mathcal{M}.
  \end{split}
  \end{equation}
  Multiple definitions of the constraint violation measure $\nu$ can be found in the literature, and the choice of which definition to use is essentially left to the search algorithm. EA commonly use $\nu$ to create penalty functions, to derive appropriate repair terms, or to rank infeasible candidate solutions. 
  A popular method to calculate the constraint violation $\nu({\bm{y}})$ of the parameter vector $\bm{y}$ is
  \begin{equation}
   \nu({\bm{y}}) = \sum_{i=1}^{l} G_i(\bm{y}) + \sum_{j=1}^k H_j(\bm{y}),
   \label{violation}
  \end{equation}
  with functions $G_i(\bm{y})$ and $H_j(\bm{y})$ defined by
  \begin{equation}
    G_i(\bm{y}) \coloneqq \max\left(0,g_i(\bm{y})\right),
  \end{equation}
  and
  \begin{equation}
    		  H_j(\bm{y})\coloneqq \left\{ \begin{matrix} \lvert h_j(\bm{y})\rvert, &\textrm{if} \:\: \lvert h_j(\bm{y})\rvert-\delta > 0\\
							      0, &\textrm{if} \:\: \lvert h_j(\bm{y})\rvert-\delta \leq 0\\ 
							 \end{matrix} \right. .
							\label{eqcon}
  \end{equation}
  In contrast to classical deterministic solvers, equality constraints cause real difficulties for meta-heuristics like EAs.
  In order to enable EA to satisfy the equality constraints at least up to a fair degree, Eq.~\eqref{eqcon} introduces the error margin $\delta$.
  Hence, parameter vectors that realize smaller deviations than $\delta$ are considered feasible. The explicit choice of $\delta$ may differ with each benchmark specification, see Sec.~\ref{cec} and Sec.~\ref{coco}.
  
  Having obtained a notion of feasibility and infeasibility of candidate solutions allows for the introduction of a corresponding order relation. 
  Such order relations permit the comparison of both feasible and infeasible candidate solutions. A commonly used order relation in the field of constrained optimization is the \emph{lexicographic ordering} $\preceq_\textrm{lex}$ which is defined in a very intuitive way. 
  Two solutions are compared at a time according to the following criteria:
	  \begin{itemize}
	    \itemsep0pt
	    \item Any feasible solution is preferred to an infeasible solution.
	    \item Among two feasible solutions, the one having the better objective function value is considered superior.
	    \item Two infeasible solutions are ranked according to their constraint violation value (the lower the better).
	   \end{itemize}
  In mathematical form, this order relation reads
  \begin{equation}
  \label{lexo}
      \bm{y} \preceq_\textrm{lex} \bm{z} \Leftrightarrow \left\{ \begin{matrix}
                               f(\bm{y}) \leq f(\bm{z}), & \textrm{if }\quad \nu(\bm{y}) = \nu(\bm{z}) = 0,\\
			       f(\bm{y}) \leq f(\bm{z}), & \textrm{if }\quad \nu(\bm{y}) =\nu(\bm{z}),\qquad \\
                               \nu(\bm{y}) < \nu(\bm{z}), & \textrm{else}.
                         \end{matrix}  \right.
  \end{equation}
  Introduced in~\cite{DEB2000}, the concept of the lexicographic ordering $\preceq_\textrm{lex}$ is also referred to as \emph{superiority of feasible solutions}. The presented order relation is commonly used for the ranking of algorithm realizations in the CEC benchmarks~\cite{CEC2006,CEC2010,CEC2017}.

 \section{Principles for EA benchmarks on constrained optimization tasks}
  \label{sec3}
  Having introduced the general problem formulation in Sec.~\ref{problem}, this section is concerned with the collection of requirements and preferable features that have to be taken into account when creating a credible 
  benchmark problem (or even framework) for constrained optimization. To this end, the already established principles used in current benchmark sets for EA are considered. Additional thoughts with respect to benchmarking guidelines~\cite{Matott2012}, experimental rigor~\cite{johnson2002}, and the presentation style~\cite{MoreWild2009} of results obtained are appended.
  
  The section is divided into three parts: the fundamental principles of the test environment, the design of adequate experiments, and the reporting of test results. Overlaps of these concepts cannot entirely be avoided. 
  
  In many cases, it is not possible to give a final recommendation of the best practice. Hence, it is not within the scope of this article to 
  provide definitive answers to these questions, but rather to create categories that allow a comparative study of distinct benchmark environments.
  Ultimately, benchmarking suites are designed with respect to various aspects of a given problem domain and certain design decisions. 
  Hence, it is the responsibility of the benchmark designers to demand the compliance of tested algorithms with these predefined benchmark principles.
  
  Taking into account publications that report on benchmarking results, ignorance of some of these principles is frequently observed.\footnote{Note, that the present paper refrains from citing bad examples.} Hence, this survey may also serve as a (by no means exhaustive) checklist to support authors and/or reviewers of such papers.
  
  \subsection{Fundamental principles of a test environment}
    \label{foundations}
    Each set of benchmark problems should ensure reproducibility of the results obtained by a specific algorithm as well as the comparability of outcome generated by other strategies. Accordingly, this subsection proposes guidelines necessary to provide a common basis for algorithm benchmarking on constrained optimization environments.
      \paragraph{Problem domain and documentation} A benchmark suite that covers all conceivable features of constrained optimization problems and their combinations appears unmanageable. Hence, it is recommended that a benchmark design systematically focuses on a specific problem subdomain instead of collecting a vast amount of arbitrary problem definitions.
      
	Well-developed benchmarking environments are supposed to guide the user through the benchmarking process.
	Users should receive clear instructions regarding the correct use of the benchmark environment, its working principles, the related benchmarking conventions, and the required reporting style. This calls for the clear definition and documentation of the related way of proceeding.
	
	 \paragraph{Problem publicity} It is to some degree necessary to decide whether the analytical description of a single problem instance 
	is openly available or whether it is generated at random. The first case allows the user to obtain a notion of the problem complexity. 
	Further, it facilitates the incorporation of real-world problems into the test problem collection. On the other hand, fixed problem statements in analytical form embrace the possibility of hand-tuning algorithm parameters for specific constrained problems or even cheating by exploiting analytical information. 
	
	Such issues can be partly circumvented by generating individual instances of a fixed constrained optimization problem at random. This involves the implementation of an elaborated test-case generator. Due to the complexity of instantiation of real-world applications, this comes with the need for designing suitable artificial test problems. 
	According to~\cite{Matott2012}, the user should not at all be involved in the evaluation of the constrained function. To this end, the benchmark collection would need to provide an easily and freely accessible software environment that offers well-defined input/output specifications.
	The availability of interfaces to multiple programming languages would additionally support the usability of such a benchmark suite.
	
      \paragraph{Function evaluations} It is imperative to provide a clear policy of how to count objective function evaluations and constraint evaluations, respectively. A first option is to interpret the evaluation of the whole constrained function, i.e. the evaluation of the objective function as well as all related constraints, as one single function evaluation. This is essentially equal to just counting objective function evaluations. Another possibility is to count the objective function evaluations as well as the constraint evaluations separately. In this case, the question remains whether to think of the constraints as a single vector-valued function that returns all constraint values at a single evaluation, or as multiple real-valued functions that account for even more function evaluations. Distinguishing between inequality and equality constraints may also represent an option. 
      More accurate counting may result in improved explanatory power, e.g. the separation of objective function and constraint evaluations allows to draw conclusions about the number of constraints inside a black-box constrained function.
      
      The finest-grained approach would be accounting the objective function and all real-valued constrained functions separately. By proper aggregation, this would still allow to use recent presentation styles (refer to Sec.~\ref{cec} and Sec.~\ref{coco}). It might further reveal insights into algorithm working principles on specific problems and with respect to different constraint types. On the other hand, depending on the constrained problem definition, the detailed information may also be used for algorithm comparison. 
      For example, given two algorithms $A$ and $B$ that show similar performance with respect to solution quality after an equal number of objective function evaluations. Observing that $A$ needs considerably less evaluations than $B$ on just one single constraint function would potentially render $A$ more preferable.
      Of course, this somehow depends on the aims and the application area of the algorithm developer.
	  
      \paragraph{Box-constraints}
	\label{boxconstr}
	A recommendation for the treatment of box-constraints needs to be stated to ensure reproducibility and comparability of the algorithm results. 
	
	According to~\cite{LiaoMMS2014}, its absence may have significant implications on the comparability of algorithm results.
	In the respective paper, it was pointed out that different box-constraint handling interpretations can produce dissimilar outcomes even for a single algorithm. The study distinguished three box-constraint scenarios:
	\begin{itemize}
	  \itemsep0pt
	  \item[(S1)] unrelaxable box-constraints,
	  \item[(S2)] relaxable box-constraints, and 
	  \item[(S3)] no box constraints at all.
	\end{itemize}
	While scenario (S3) is self-explanatory, the box-constraints are defined and enforced at any stage of the search process in situation (S1). Candidate solutions outside the box are considered invalid and thus have to be repaired or discarded.
	In case of (S2), box-constraints are specified, but only enforced for the final candidate solutions. That is, infeasible candidate solutions outside the box may be used to drive the search.
	It was shown in~\cite{LiaoMMS2014} that algorithms were sometimes able to find solutions of better quality when facing situation (S2) or (S3) instead of (S1), and even if the global optimizer was not located on the boundary of the specified box $\mathcal{S}$.
	
	In order to avoid inconsistencies, various options come to mind. 
	First, the box-constrained treatment can be completely eliminated if the admissible intervals of the parameter vector components $y_i$ are directly included in the inequality constraints $g_i(\bm{y})$. In case of one specific lower and upper bound for each parameter vector component, the number of inequality constraints increases by $2N$. 
	Regarding high dimensional problems, one can think of situations where this can potentially blow the problem complexity out of proportion.
	Inducing that most algorithms would have to be adapted, this approach would limit the usability of such a benchmark problem.  
	However, the least invasive option is giving permission to apply the individual box-constraint handling techniques of choice. 
	This clearly comes with the need for a proper reporting of its corresponding modus operandi.

 \subsection{Experimental design}
      \label{experiments}
	 The experimental design of a benchmark testbed is supposed to properly reflect the characteristics of the chosen problem (sub)domain. This requires the unambiguous description of the constrained test problems, initialization practices, as well as appropriate quality indicators.
	 The benchmark problems are expected to be efficiently implemented in order to speed up the experiments. 
	 Moreover, the following subjects have to be adopted in the design process.
  	 
   \paragraph{Initialization} 
	   Differences with respect to the initialization parameter vectors are present. These have varying implications on the applicability of certain optimizers.
	   A benchmark problem might either provide a feasible initial candidate solution, supply a subset of not necessarily feasible parameter vectors (e.g. by specifying unrelaxable box-constraints), or give no assistance at all. In case that no feasible solution is given, algorithms that rely on initially feasible solutions essentially have to priorly solve a constraint satisfaction problem before the original constrained optimization problem is tackled. This can significantly impair their performance and would complicate the comparability of such approaches with strategies that do not assume the existence of a feasible solution.    
	   
      \paragraph{Precision}
	  Considering randomized algorithms, a test environment needs to make assumptions on the termination precision and reasonable error margins for constraint satisfaction. The latter is particularly important in the context of equality constraints because it is otherwise highly improbable to find feasible candidate solutions. 
	  Further, a statement on the required precision of reported statistics appears necessary to ensure an appropriate ranking of two distinct algorithms on a single constrained function. For example, assuming two algorithms A and B both reliably approach the optimal objective function value of zero on the same constrained function. While A realizes a mean function value of $10^{-10}$ in multiple, independent runs, B achieves a mean value of $10^{-11}$. Ranking algorithm B better than A based only on the observed mean values is quite questionable in this scenario.
	  Considering precisions below the floating point accuracy also appears misguided.
	  
	  Actually, although it is commonly done, the consideration of relative precisions (or absolute precisions in the case of $f(\bm{y}^*)= 0$) of order $10^{-6}$ or even smaller does not always reflect the needs of real-world optimization problems. That is, at some point of the search process the effort to realize very small improvements might be expendable from a practical point of view.
	   	
      \paragraph{Constrained problems}
	  A sufficient number of profound constrained optimization problems suitable to represent the chosen problem domain need to be appointed. The problems might either be automatically generated or collected from test problem collections. Each problem needs to be specified in the manner of~\eqref{cop}. That is, objective function, constraint functions, and box-constraints have to be well-defined. In case that this information is not made public, a black-box framework has to be developed that supplies the objective function value and at least an indicator of constraint satisfaction (or violation) to the solver.
	  
	  Taking into account that current algorithms have to deal  with continuously increasing problem complexity, the constrained functions are ideally designed in a scalable fashion~\cite{Whitley1996}. Scalability with respect to the search space dimension, and also the number of constraint functions, permits an understanding of the inherent problem complexity. It further allows assessing these factors of influence on the algorithm performance.
	  In this regard, the creation of artificial test problems represents a much easier way to generate constrained test functions.
	  On the downside, such test problems are usually easier to solve than real-world problems. 
	  However, real-world problems are hardly scalable as they often state a purpose-built mathematical representation of a certain application. Modifications in terms of dimension or constraint numbers may result in a change of the problem structure.
	  Further, the design of constrained test functions should incorporate characteristics that are commonly observed in real-world situations. This way, algorithmic ideas that proved themselves successful on the benchmark suite can be transferred to corresponding real-world applications with partly similar characteristics.
		  
	  Building clusters of constrained problems with similar features facilitates insight into the algorithm performance on each of the problem subgroups. It further supports the decision whether an algorithmic idea is useful when dealing with specific real-world applications of a certain characteristic~\cite{Mersmann2015}. For example, regarding a practical application that involves satisfying a great number of constraints, algorithms that have been observed to perform well on test problem subgroups with similar features are of interest. These are usually expected to be better suited than the collectively best algorithm which ultimately might represent a compromise over all benchmark problems.
	  	  
	  Moreover, the design of problem instances preferably should exclude biases towards certain algorithm classes.	
	  To this end, problem formulations aligned in the Cartesian axes should be avoided. Further, problems whose optimum is located on the boundary of the box $\mathcal{S}$ may exhibit the tendency to favor EA that use specific box-constraint handling techniques.
	  Such issues may be bypassed by considering different instances of a problem, e.g. by introducing small modifications with respect to the orientation of the feasible region or the location of the optimum (see Sec.~\ref{problem}).
	  The creation of new instances is usually simpler for theoretically derived constrained functions.
	  Real-world problems determined by specific application cases usually have a rather rigid formal representation without any information about the optimum.
	
	\paragraph{Order relation} 
	Benchmark environments that compare algorithms on the basis of solution quality need a consistent order relation 
	for ranking the provided candidate solution realizations. To this end the order relation should be able to deal with feasible and infeasible candidate solutions. A commonly used approach is the so-called \emph{superiority of feasible solutions}~\cite{DEB2000} which is recapped in Eq.~\eqref{lexo}. Benchmarking environments might take into account different ordering instructions. However, these need to be motivated convincingly.

	 \paragraph{Quality indicators}  
	    Multiple aspects of algorithm performance have to be covered by the experimental design~\cite{Matott2012,johnson2002}. 
	The benchmark environment has to use a number of well-defined quality indicators that are computed in the experiments.
	The quality indicators reflect the suitability of a respective algorithm for a specific constrained function, a subgroup of  constrained function, and the whole problem collection. Moreover, the quality indicators build the basis for algorithm comparison.
	That is, the benchmarks essentially need to introduce measures of \emph{effectiveness}, \emph{efficiency} and \emph{variability}.
	A high \emph{effectiveness} of an algorithm refers to its ability to realize solutions close the best-known or optimal solution of a problem. On the other hand, an \emph{efficiency} measure accounts for the number of resources (e.g. function evaluations or time) consumed for computing high-quality solutions. Further, a measure of \emph{variability} quantifies the reliability of an algorithm to realize equally good candidate solutions in multiple independent runs.
There exist multiple ways to define such indicators. Hence, it is left to the benchmark designers to choose the most appropriate measures of algorithm performance for the corresponding problems. 	 
		       
	To obtain the quantity of variability, benchmarking of randomized algorithms involves running multiple independent algorithmic runs on the same problem instances. The appropriate number of repetitions is connected to the choice of quality indicators~\cite{MoreWild2009}. In order to obtain reasonable statistics a minimum number of 10 to 25 algorithm runs is usually recommended.

    \paragraph{Termination}
	 A benchmark collection might determine strict rules on the termination conditions for participating algorithms, e.g. a fixed budget of function evaluations.
	 Another approach would be to set multiple targets for an optimization strategy. Termination takes place after hitting the final target. By measuring the number of functions evaluations needed to reach a specified target a notion of algorithm speed can additionally be established. However, introducing targets assumes knowledge about the optimal function values of the constrained problems.

  \subsection{Reporting}
    \label{reporting}
    
	  This section takes into account useful principles that support a reproducible and comprehensible presentation of obtained algorithm results.
	  Further, it is concerned with the aspect of algorithm comparison and mentions the need for encouraging algorithm developers to thoroughly report algorithmic details.

   \paragraph{Newsworthiness and presentation} 
	 To ensure that meaningful results are generated, the benchmarking environment can support the user by providing a performance baseline. 
	 Such a baseline may represent performance results obtained by application of comparable algorithms for constrained optimization. 
	 If a collection of algorithm results is not present, even the performance results of random search can be considered useful. 
	 Such information is necessary to  realize whether the benchmarked algorithm is, in fact, superior for a number of problems. 
	  This way, publications with respect to already dominated algorithmic ideas can be avoided.  
	  
	 The performance results have to be presented in informative ways to support the interpretation of the individual algorithmic behavior. 
	 This is preferably realized by stipulating a presentation style that uses a combination of tables and figures.
	 By providing aggregated algorithm results for the complete benchmark collection as well as for predefined constrained function subgroups, 
	 the benchmark suite allows for establishing a connection between a tested algorithm and suitably constrained problems that it can solve.

    \paragraph{Ranking of algorithms}
	Alongside with the presentation of individual algorithm performance, it is the purpose of a benchmark environment to answer the question which algorithm is best suited for solving (a subset of) the benchmark problems.
	The comparability of the algorithmic results is ensured by defining an appropriate ranking procedure. 
	
	Regarding constrained benchmarking functions, the comparability of algorithms results is in need of an ordering approach that is able to distinguish between feasible and infeasible realizations of the obtained quality indicators. A suitable representation of such an order relation is provided by the lexicographic ordering that has been defined in the context of Sec.~\ref{problem}. 
	By introducing an order of priority to multiple quality indicators, the lexicographic ordering can be analogously defined to determine a proper algorithm ranking. The question which quality indicators to use for ranking competing algorithms involves a certain degree of subjectivity.
	For that reason, it is recommended in~\cite{Mersmann2015} to make use of consensus rankings which comprise more than one order relation. 
	This way, a consensus ranking allows computing an appropriate algorithm ranking over the whole benchmark suite, or subsets of constrained problems, respectively.
								
	In order to decide whether comparably small performance differences can be considered significant, the algorithm comparison usually benefits from factoring in statistical hypothesis testing. Being less restrictive than parametric approaches and requiring smaller sample sizes, non-parametric tests are usually recommended when testing EA realizations for statistical significance~\cite{Garcia2009}. However, statistical and practical significance are not necessarily equivalent and a well-established graphical representation of the algorithm results may suffice~\cite{rardin2001experimental}.
								
    \paragraph{Algorithm description}
		 When providing benchmark results, it should be mandatory to require a proper characterization of a tested algorithm. Such a description includes the detailed motivation of prior investigations and a comprehensive description of the implemented algorithmic ideas. Further, an exact pseudo-code representation is desirable to illustrate the working principles. 
		 Among others, this includes a specification of box-constraint handling techniques, or the use of (approximated) gradient information, respectively.
		 All algorithm specific strategy parameters need to be reported together with an explanation of their impact on the algorithm performance at best. 
		 		
    \paragraph{PC configuration} 
	  When it comes to measuring the computational running time of an algorithm, the users of benchmark collections should be required to report on the complete PC configuration. This includes detailed information about the processor architecture, memory, operating system, and the programming language, confer~\cite{CEC2017}. 
	  The use of performance benchmarks to calibrate algorithm speed is also recommended in order to obtain a perception of the system-depending performance. This way algorithm comparability can be maintained over long periods of time~\cite{johnson2002}.
	  
    \paragraph{Runtime and algorithm complexity}
	  Algorithm efficiency can be assessed by accounting the number of resources needed to reach a given high-quality solution.
	  
	  To this end, the CPU time (or wall-clock time) needed for a predifed number of elementary operations can be determined. 
	  The consumed time provides an estimate of the algorithm complexity.
	  In order to ensure comparable results, baseline measurements are necessary. However, measureing algorithm efficiency by means of CPU time is machine-dependent and comes with reduced reproducability (if performance benchmarks are omitted, see {\em PC configuration} above).
	  
	  According to~\cite{hansen2016perf}, a machine-independent performance criterion suitable for direct search algorithms is the algorithm runtime in terms of the number of function evaluations executed. That is, the measurement of CPU time can be regarded irrelevant in the context of derivative-free optimization. This approach assumes the availability of well-defined algorithm targets, e.g. the knowledge about the optimal solution of a constrained function that has to be approached with reasonable accuracy. 
	  Algorithm efficiency can then be identified with the number of function evaluations consumed until the (final) target is reached.
	  
	  Further, benchmark suites may concentrate on the computation of different indicators like mean or median solution quality.
	  Such studies may argue that their focus is limited on the effectiveness of the algorithms and that runtime can be neglected in this context.	  
     	  Yet, regardless of the primary goals of a benchmark set, the algorithm running time should be reported~\cite{johnson2002}.
     	  It can be used to indicate algorithm complexity, i.e. running time trade-offs that are related to increased solution quality and vice versa.
     	  Further, it provides a notion of the computational effort for reproducing the reported results and 
     	  may provide useful information for assessing parallelization attempts.
     	  
	  Anyway, plain instructions for computing the algorithm speed have to be provided. 
	  This is achieved by indicating whether the calculations are performed for only one exemplary algorithm run or whether it considers all repetitions.
	  Further, the running time may cover all preprocessing and initialization steps, or it might only focus on the main loop of the considered algorithm.
	  Ideally, the complete algorithms time should be measured and reported relative to reproducible performance benchmarks.

\section{The CEC competition on constrained real-parameter optimization}
\label{cec}

   The test function sets defined in the context of the IEEE Congress on Evolutionary Computation (CEC) competitions on single objective constrained real-parameter optimization are arguably the most common test collections for benchmarking randomized search algorithms.
The CEC competitions have been organized in 2006~\cite{CEC2006}, 2010~\cite{CEC2010}, and 2017~\cite{CEC2017}. Each of these competitions introduced a specific set of constrained test problems in the line with~\eqref{cop}. The test functions sets are supported with a policy for the computation of comprehensive performance indicators and for reporting algorithm results.

The remainder of this section is concerned with reviewing the benchmarking conventions associated with the mentioned CEC benchmark environments as well as their characteristic features. To this end, the benchmark definitions are examined by taking into account three different aspects: the basic benchmarking conventions, the experimental setup, and the reporting of algorithm results. A summary of important features of the three constrained benchmark environments is provided in Table~\ref{tab02}.
 
\subsection{Benchmarking conventions}
\label{cec_foundations}

The CEC2006 benchmarks\footnote{Note that the present paper refers to the constrained test problem set specified for the competition in year 2006 as \emph{CEC2006 benchmarks}. The denotations \emph{CEC2010 benchmarks} and \emph{CEC2017 benchmarks} have to be understood in analogous manner.} build a test environment of 24 distinct constrained functions with various features. The first 11 constrained problems (\textrm{p01} to \textrm{p11}) were originally collected in~\cite{MichalewiczS1996}, problems \textrm{p12} and \textrm{p13} are taken from \cite{Michalewicz1995,Koziel1999}, problems \textrm{p21} and \textrm{p22} can be traced back to heat exchange network applications~\cite{epperly1995global}, \textrm{p23} was suggested in~\cite{Xia96}, and \textrm{p24} can be found in~\cite{floudas1999handbook}. For the remaining test problems (\textrm{p14} to \textrm{p20}) it is referred to~\cite{himmelblau1972applied}.

The succeeding benchmark definitions for CEC2010~\cite{CEC2010} introduced 18 new constrained benchmark problems. Yet, the origin of the corresponding constrained functions is not easily comprehensible. Only one constrained function was adopted from the CEC2006 benchmarks. The benchmark set introduced variations of 8 distinct objective functions that differ with respect to the application of parameter translations and/or rotations. Further variations are obtained by introduction of different number and types of constraint functions. Some objective and constraint functions can be attributed to a collection of unconstrained problems~\cite{CEC2005,schwefel1995evolution}, e.g. the Rosenbrock function, the Griewank function, and the Weierstrass function. Other function definitions were obtained by use of the test-case generator proposed in~\cite{Michalewicz2000}.
However, being defined in scalable from with respect to the search space dimension, the constrained test problems have to be solved in dimension $N=10$ and $N=30$.
 \begin{table}[t] 
  \centering
  \renewcommand{\arraystretch}{1.04}
    \begin{tabular}{p{5.5cm}p{1.5cm}p{1.7cm}p{1.7cm}}
 	Benchmark name & CEC2006  & CEC2010 & CEC2017 \\ \hline \hline
 	Minimal $N$ & $2$& $10$& $10$ \\[0.5ex]
 	Maximal $N$ & $24$ & $30$& $100$\\[0.5ex]  \hline
 	Number of constrained functions  & $24$ & $36$ & $112$ \\[0.5ex]
 	Number of distinct obj. functions & $ 23$  & $8 $ & $15$ \\[0.5ex]
 	Minimal number of constraints & $1$& $1$& $1$ \\[0.5ex]
 	Maximal number of constraints & $38$ & $4$& $6$\\[0.5ex]
 	Avg. number of constraints & $7.0$ & $2.1$ &  $2.2$ \\[0.5ex]\hline
	Scalable problems included   & no & yes  & yes \\[0.5ex]
	Budget of function evaluations  & $5\cdot 10^5$ & $2\cdot 10^4\cdot N$ & $2 \cdot 10^4\cdot N$ \\[0.5ex]
 	Number of fully separable problems (objective and constraints) & $6$ & $4$ & $20$\\[0.5ex] \hline
	Avg. size of $\rho = \mathcal{M}/\mathcal{S}$ & $11.3\%$ & $8.9\%$ & $3.4\%$ \\[0.5ex]
	Number of problems with $\rho > 10^{-3}$ & $5$ & $10$ & $12$ \\
     \end{tabular}
    \caption{Characteristic features of the CEC benchmark sets for single objective constrained real-parameter optimization.} 
    \label{tab02}
 \end{table}
 
Considering even larger search space dimensions ($N=10$, $N=30$, $N=50$, and $N=100$), a novel collection of 28 benchmark problems was created for the CEC2017 competition~\cite{CEC2017}. The 2017 constrained function definitions are designed by taking new combinations of the building blocks provided in~\cite{CEC2005,schwefel1995evolution,Michalewicz2000}. However, some overlaps do exist.
It is claimed that the CEC2006 benchmarks and the CEC2010 have been successfully solved~\cite{CEC2017}. Yet, the older CEC testbeds are still very popular 
for benchmarking direct search algorithms and particularly Evolutionary Algorithms, e.g.~\cite{GONG2014884}.
In contrast, the CEC2017 problem definitions are reutilized for the CEC competition on single objective constrained real-parameter optimization taking place during the IEEE World Congress on Computational Intelligence (WCCI) in 2018.

The constrained function definitions are fully presented in the corresponding technical reports. Yet, some constrained problems lack a description of the translation vectors and rotation matrices. These can only be understood by taking into account their implementations. The corresponding code is maintained on the respective website of the competition organizers~\cite{websource01}. It is openly available in the programming languages C and MATLAB. 

The consecutive development from CEC2006 towards the CEC2017 benchmarks is not entirely motivated in the corresponding technical reports. 
Modifications with respect to performance indicators or algorithm ranking approaches are not entirely transparent.
The documentation sometimes leaves room for interpretations by inexact instructions.

All three technical reports~\cite{CEC2006,CEC2010,CEC2017} of the constrained CEC benchmark collections demand to identify the evaluation of the whole constrained function as one single function evaluation. 
That is, each constrained function evaluation consumes one function evaluation of the predefined budget regardless of whether the objective function value or only some constraint function values associated with a single candidate solution are of interest.
The use of gradient information is only applicable if the gradient is approximated numerically and the consumed function evaluations are properly taken into account.

The CEC competitions for constraint real-parameter optimization do not enforce the feasibility of search space parameter vectors.
In this respect, equality and inequality constraints of a constrained function~\eqref{cop} are always considered as relaxable, cf. option (S2) in Sec.~\ref{problem}. 
That is, the algorithms are allowed to move in the unconstrained search space. Each candidate solution, either feasible or infeasible, may be evaluated and used within the search process of a strategy. 
Using only relaxable (in-)equality constraints represents a reasonable design decision common for EA benchmarking (cf. Sec~\ref{coco}).
However, it should be mentioned that the permission to use infeasible solutions during the search may significantly reduce the problem complexity.

For instance, algorithms might be allowed to solely operate outside the feasible region until the optimizer is approached sufficiently close. 

A specific treatment of box-constraints is not stipulated by the CEC benchmarks. The technical reports are not clear on whether box-constraints have to be regarded relaxable (S2) or unrelaxable (S1). This ambiguity can potentially result in different approaches, and ultimately in significant performance differences~\cite{LiaoMMS2014}.
Taking into account the most successful strategies reported in CEC competitions~\cite{HuangQS06,TakahamaS06,TakahamaS10,MallipeddiS10,PolakovaT2017,TvrdikP2017} and after inspecting the related openly available source codes, up to our knowledge, all algorithms were assuming situation (S1) as introduced in Sec.~\ref{problem}. 
Albeit reporting the full algorithm can be considered scientific standard, yet some papers miss out on giving such information. 
Further, the mechanisms to treat box constraint violations may vary. To ensure the reproducibility of the benchmark results, the testbeds have to explicitly demand a statement on the box-constraint handling techniques used by an algorithm.

For the computation of the quality indicators (see Sec.~\ref{cec_report}), the CEC framework sorts the algorithm realizations of 25 independent runs on the basis of the lexicographic ordering relation introduced in~\eqref{lexo}. That is, feasible solutions are ranked based on their objective function values. They always dominate infeasible solutions which are distinguished with respect to the related magnitude of their mean constraint violation (see Sec.~\ref{cec_exp}, Eq.~\eqref{meanvio}).

\subsection{Experimental design}
\label{cec_exp}
The CEC competitions on constrained real-parameter optimization do not provide an initially feasible region or candidate solution. Instead individual box-constraints are specified for each constrained problem and algorithms are supposed to randomly sample a starting point or an initial population inside of the set $\mathcal{S}$  of problem~\eqref{cop}.

Hence, the feasibility of initial candidate solutions is not ensured. In order to be competitive on the CEC benchmarks, algorithms need to be able to deal with infeasible solutions. This is affirmed when considering the size of the feasible region $\mathcal{M}$ relative to $\mathcal{S}$, i.e. the parameter $\rho$ (cf. Eq.~\eqref{relsize}). Looking at Table~\ref{tab02}, the average $\rho$ value was reduced over the years. Whereas the ratio of constrained problems with a feasible region greater than $0.1\%$ was $7/24$ in 2006, this number dropped to $8/36$ in 2010, and even further to $12/112$ for constrained functions specified in 2017\footnote{Only, the reports on the CEC2006 and CEC2010 reported on the $\rho$ values. To ensure comparability, the $\rho$ values of the constrained CEC benchmark problems have been reevaluated by use of the method presented in~\cite{Koziel1999}.}.
In consequence, the benchmark sets contain many problems with very small $\rho$ values. 
The feasible region of some problems only consists of few disjoint areas in the parameter space. For these constrained functions it is of course very difficult to generate feasible solutions in the first place. In this regard, algorithms that initially (or completely) rely on a feasible solution appear ill-equipped for many constrained functions in these benchmark sets.



Regarding the CEC2006 competition, the detailed benchmark function specifications can be found in the technical report~\cite{CEC2006}. The benchmark set consists of $24$ constrained functions of varying search space dimensions between $N=2$ and $N=24$.
The given constrained functions are fixed in terms of the problem dimension and the number of constraints.
Each objective function is restricted by in between $1$ and $36$ linear and non-linear (in-)equality constraints, refer to Table~\ref{tab02}.
The optimal solution, or at least the best-known solution, is provided for each constrained function.

The 2006 benchmarks include $6$ fully separable constrained functions. Refraining from the use of parameter vector rotations, the benchmarks enclose a potential bias towards strategies that search predominantly along the coordinate axes of the search space~\cite{SuttonLW2007}. 
In this regard, the CEC2006 benchmarks favor algorithms that use coordinate-wise search or differences of obtained candidate solutions, e.g. Coordinate Search or Differential Evolution variants. 

The benchmark definitions of the CEC2010 competition~\cite{CEC2010} can be considered a refinement with respect to this issue.
As mentioned above, the constrained problems of CEC2010 can be affiliated to different sources~\cite{CEC2005, schwefel1995evolution} and are partly designed by use of the test-case generator~\cite{Michalewicz2000}.
The 2010 competition included $36$ constrained functions in dimensions $N=10$, and $N=30$, respectively.
The formulation of scalable constrained functions allows for conclusions with respect to an algorithm's ability to deal with growing search space dimensions.
The mentioned bias towards coordinate search and separability was partly resolved by application of predefined search space rotations.
Each objective function is accompanied with from $1$ to $4$ constraint functions. Hence, the average number of constraints per constrained function drops from $7$ in 2006 to about $2.1$ in 2010.
In this respect, the CEC2010 competition problems represent a fresh start instead of being a progression of the CEC2006 problem definitions. 
The question to what extent the small number of constraints can actually cover real-world problem aspects remains.
Moreover, best-known solutions to the benchmark problems are no longer reported. 
This impedes gathering information about the effectiveness of an algorithm.

Still, $4$ out of $36$ problems are fully separable and do not apply any rotations to the parameter vectors.
While the formal description of those transformations is not satisfactorily explained in the technical report, 
it is deposited in the corresponding competition source code~\cite{websource01}. 
There, the transformations are deterministically specified, and different, for each individual constrained function. 

Having a look at the CEC2017 competition, the constrained function definitions are quite similar to its predecessor competition.
The corresponding technical report~\cite{CEC2017} states $28$ scalable constrained optimization problems essentially attributable to the same sources of the CEC2010 benchmarks. 
The latest CEC collection considers not only a larger number of problems but also larger search space dimensions: $N=10$, $N=30$, $N=50$, and $N=100$. 
In total, the competition comprises $112$ constrained functions. The number of constraints is between $1$ and $6$, i.e. the average number of constraints per problem is 
comparable to the CEC2010 benchmarks (refer to Table~\ref{tab02}).
Similarly to the 2010 version, information on optimal parameter vectors or function values is omitted. 
Among these problem definitions, $16$ out of $112$ constrained functions are separable. 
To this end, a small bias towards strategies that predominantly search parallel to the Cartesian axis of the search space cannot be fully excluded.

The CEC benchmark environments do not establish subgroups of constrained problems.
That is, results obtained by an algorithm can hardly be identified with a certain problem characteristic. 
Although, the CEC2017 collection would allow for a rough categorizations. 
For example, the constrained problems $\left(\textrm{p01}| \textrm{p02}| \textrm{p03}\right)$, $\left(\textrm{p08}| \textrm{p09}| \textrm{p10}\right)$, and $\left(\textrm{p05}| \textrm{p13}| \textrm{p22}\right)$ respectively, share the same objective function but differ in the number and type of their constraint functions.
Problem classes that address the number or type of the constraints would also be conceivable.
This would be useful for extracting additional information about the applicability of algorithmic ideas to such problem classes.

All CEC benchmark sets share the definition of a feasible solution introduced in Sec.~\ref{problem}. 
Due to the issue of enforcing the generation of candidate solutions that exactly satisfy the equality constraints, the error margin of $\epsilon =10^{-4}$ is used in all three competitions.

Every algorithm has to perform $25$ independent runs on a single instance of each constrained optimization problem.
In each run, the best result so far $\bm{y}_\textrm{bsf}$ is monitored at three distinct points of the search process, i.e. after 10\%, after 50\%, and after 100\% of the assigned function evaluation budget have been consumed.\footnote{Notice that, for the CEC2006 benchmarks the same measurements had to be collected after $1\%$, $10\%$ and $100\%$ of the evaluation budget.} 
To this end, an algorithm is required to report the best so far objective function value $f(\bm{y}_\textrm{bsf})$, the corresponding mean constraint violation $\nu(\bm{y}_\textrm{bsf})$, as well as the triplet $\bm{c}$ (see Table~\ref{tab03}). 
The mean constraint violation $\bar{\nu}({\bm{y}})$ of a candidate solution $\bm{y}$ is determined as
  \begin{equation}
  \label{meanvio}
   \bar{\nu}({\bm{y}}) = \cfrac{\nu({\bm{y}})}{m},
  \end{equation}
where $m$ is the aggregated number of equality and inequality constraints of problem~\eqref{cop}. 
Note that the constraint violation $\nu({\bm{y}})$ is obtained according to Eq.~\eqref{violation}.
The term $\bm{c}$ specifies the number of violated constraints with violation greater than $10^{0}$, $10^{-2}$, and $10^{-4}$ , respectively.

The results of these $25$ runs are then used to compute statistics for algorithm evaluation and comparison. 
In order to sort the realized candidate solutions, the CEC benchmarks introduce a lexicographic ordering with respect to $f$ and $\bar{\nu}$. 
That is, two candidate solutions $\bm{y}$ and $\bm{z}$ are sorted according to
  \begin{equation}
  \label{lexo2}
      \bm{y} \preceq_\textrm{lex} \bm{z} \Leftrightarrow \left\{ \begin{matrix}
                               f(\bm{y}) \leq f(\bm{z}), & \textrm{if }\quad \bar{\nu}(\bm{y}) = \bar{\nu}(\bm{z}) = 0,\\
			       f(\bm{y}) \leq f(\bm{z}), & \textrm{if }\quad \bar{\nu}(\bm{y}) =\bar{\nu}(\bm{z}),\qquad \\
                               \bar{\nu}(\bm{y}) < \bar{\nu}(\bm{z}), & \textrm{else}.
                         \end{matrix}  \right.
  \end{equation}
  Note that Eq.~\eqref{lexo2} is defined analogously to the order relation~\eqref{lexo}, but makes use of Eq.~\eqref{meanvio} instead of Eq.~\eqref{violation}.
A comprehensive list of the utilized quality indicators is provided in Table~\ref{tab03}.
  \begin{table}[!htbp]
  \centering
  \renewcommand{\arraystretch}{1.1}
    \begin{tabular}{lp{6cm}ccc}
       \textbf{Notation} & \textbf{Description} & \textbf{2006}  & \textbf{2010} & \textbf{2017}\\ \hline 
	  Best  & The objective function value $f(\bm{y}_\textrm{best})$ corresponding to the best found solution $\bm{y}_\textrm{best}$ in 25 independent algorithm runs with respect to Eq.~\eqref{lexo2}.  & + & + & + \\[1ex]
	  Median & The objective function value $f(\bm{y}_\textrm{median})$ associated with the median solution $\bm{y}_\textrm{median}$ of the 25 algorithm realizations according to Eq.~\eqref{lexo2}. & + & + & + \\[1ex]
	  $\bm{c}$ & A vector containing the number of constraints with violation greater than $10^{0}$, $10^{-2}$, and $10^{-4}$ associated with the median solution.  &+ &+ &+\\[1ex]
	  $\bar{\nu}$ & The mean constraint violation value $\bar{\nu}(\bm{y}_\textrm{median})$ associated with the median solution $\bm{y}_\textrm{median}$, refer to Eq.\eqref{meanvio}.  &+ &+ &+\\[1ex] 
	 Mean & The mean objective function value according to the 25 independent algorithm runs.   & +& +&+\\[1ex]
	 Worst & The objective function value $f(\bm{y}_\textrm{worst})$ corresponding to the worst found solution $\bm{y}_\textrm{worst}$. &+ &+ & +\\[1ex]
	 Std & The standard deviation according to the objective function values obtained in 25 runs.  &+ & +&+\\[1ex]
	$FR$ & The ratio of feasible algorithm realizations over the number of total runs.  & +&+ &+\\[1ex]
	$SR$ & The ratio of successful algorithm runs, cf.~\eqref{successRate}, over the number of total runs was computed.  & +&- &-\\[1ex]
	$SP$ & The quotient of the mean number of function evaluations consumed in successful runs and the success ratio is referred to as success performance $SP$. & +&- &-\\[1ex]
	$\overline{vio}$ & The mean constraint violation corresponding to the 25 independent algorithm runs. & -& -&+
    \end{tabular}
    \caption{Quality indicators computed for the CEC competitions on constrained real-parameter optimization. The $+/-$ markers indicate whether the respective quality indicator is used in a CEC benchmark set.} 
    \label{tab03}
  \end{table}

The CEC2006 benchmark set provided the globally optimal parameter vectors of each test problem.
 Using this information the effectiveness of an algorithm was determined in terms of the deviation $\left( f(\bm{y})-f(\bm{y}^*) \right)$ of the best-so-far solution $\bm{y}$ from the optimum $\bm{y}^*$.
It was further used to calculate the success rate ($SR$) of a specific algorithm. The success rate was defined as the ratio
of successful runs and the number of total runs. Hence, an algorithm run is considered successful if at least one feasible solution with 
  \begin{equation}
    \label{successRate}
   \left( f(\bm{y})-f(\bm{y}^*) \right) \leq 10^{-4} 
  \end{equation}
is realized. 
Note that, by distinguishing two feasible candidate solutions based on their deviation from the known optimum, the CEC2006 benchmarks use a slightly different way of proceeding than presented in~\eqref{lexo2}.

No longer having information about the global optima, the success rate was replaced with the calculation of the feasibility rate ($FR$) in the succeeding CEC competitions. $FR$ indicates the ratio of those algorithm runs that realized at least one feasible solution and the total number of algorithm runs.
 
  Regarding the termination criterion used by the CEC competitions, each constrained problem comes with a fixed budget of function evaluations.\footnote{Keeping in mind, that the CEC benchmark definitions refer to a function evaluation as one evaluation of the whole constrained function, see Sec.~\ref{cec_foundations}.} 
  Termination is required after an algorithm has entirely consumed this budget. 
  The budget of function evaluations allocated to each constrained function varies among competitions. While it is fixed to $5\cdot 10^5$ function evaluations  (regardless of the problem dimension) for CEC2006, the CEC2010 and CEC2017 collections define the budget proportional to the problem dimension $N$. That is, each algorithm is allocated a budget of $2\cdot 10^4 \cdot N$  function evaluations. Other termination criteria are not designated.

\subsection{Reporting}
\label{cec_report}
By primarily representing test problems for the CEC competitions on constrained real-parameter optimization, the corresponding technical reports do not make a statement on ensuring newsworthiness of the algorithm results. 
In order to participate in the mentioned competitions, algorithm results have to be published in a conference paper that has to pass a related review process.
Accordingly, the novelty of algorithms is reviewed in this way.
However, those authors that use the constrained CEC functions as benchmarks in a different context might need to be reminded of assessing the benefit of their algorithmic ideas. To this end, benchmark results of comparable algorithms should be supplied, e.g. results obtained by the winning strategies from earlier competitions or even by random search.
Such information is considered useful for quickly evaluating the suitability of a novel algorithm and its competitiveness for the CEC competitions.
As pointed out by~\cite{Garcia-Martinez2017} in the context of unconstrained benchmarks, the comparison of novel algorithmic ideas with diverse state-of-the-art strategies is essential to prevent the publication of already dominated results and to contribute to real progress in the respective field of research.

The final quality indicators computed for a specific algorithm have to be presented for every single constrained problem in a detailed table.
Considering that the CEC benchmarks demand information on three stages ($10\%$, $50\%$, and $100\%$) of the search process, this presentation style appears rather lengthy. 
Table~\ref{tab04} illustrates the presentation guidelines corresponding to the CEC2017 benchmarks.
\begin{table}[t]
  \centering
    \begin{tabular}{l|l|p{1cm}|p{1cm}|p{1cm}p{1cm}p{1cm}}
       Budget & Indicator & p01 & p02 & p03 & \dots & p28 \\ \hline 
	  & Best & & & & \\
	 & Median & && & \\  
	 & $\bm{c}$ && & & \\  
	 & $\bar{\nu}$&& & &  \\  
	10\% & Mean & && & \\  
	 &Worst & & && \\  
	 & Std  & && & \\  
	 & $FR$ & && & \\  
	 & $\overline{vio}$ & && &  \\ \hline
	 & Best & & & & \\[-1ex]
	50\% &\vdots & &&  \\
	 &$\overline{vio}$&&& & \\ \hline
	 &Best &&& & \\[-1ex]
	100\%&\vdots && & & \\
	 &$\overline{vio}$ &&& &\\ \hline
    \end{tabular}
    \caption{Presentation of algorithm results obtained in dimension $N$ according to the guidelines of the CEC2017 competition on constrained real-parameter optimization~\cite{CEC2017}.} 
    \label{tab04}
  \end{table}
Making use of one table per dimension, and per algorithm, leads to increasing space requirements when considering more search space dimensions.
Furthermore, drawing conclusions with respect to algorithm performance differences is made very difficult.
Additionally, not subsuming problems of similar characteristics impedes interpretations of the results.

The CEC2006 and CEC2010 benchmarks were using convergence graphs to provide a more tangible notion of algorithm performance. 
In 2006, the convergence graphs illustrated the deviation of the objective function value from the optimum $\left(f(\bm{y})-f(\bm{y}^*)\right)$ as well as the mean constraint violation $\bar{\nu}(\bm{y})$ plotted against the number of function evaluations in full-logarithmic scales. Instead of taking into account the median solution, the technical report of CEC2010 recommends illustrating the best out of 25 runs.
The idea of convergence graphs was dropped with growing table sizes for the CEC2017 competition.

The benchmark collections demand to report the configuration of the PC on which the experiments have been executed.
To this end, the \emph{operation system}, the \emph{CPU}, the \emph{memory}, the \emph{programming language} used, and the \emph{algorithm} have to be specified.
Acting this way intends to support algorithm comparability. However, a performance benchmark to calibrate a tested algorithm's efficiency on the corresponding system is not recommended. Such a performance baseline would retain comparability of algorithm results obtained on outdated systems.

With respect to algorithm reporting, the CEC related technical reports~\cite{CEC2006,CEC2010,CEC2017} require the complete description of the algorithm parameters used as well as their specific ranges.
Further, algorithm designers are demanded to present guidelines for potential parameter adjustments and estimates of the corresponding costs in terms of function evaluations.
The use of hand-tuned parameters for individual constrained functions is interdicted.

In order to give an impression of the algorithm complexity (see Sec.~\ref{reporting}), three quantities have to be presented. 
The average $T1$ of the computation time $t_1^i$ of $10^4$ evaluations, as well as $T2$, the complete computation time $t_2^i$
of a specific algorithm over all problems $i\in\{1,\dots,np\}$ of similar dimensionality
\begin{equation}
 T1 = \cfrac{\sum\nolimits_{i=1}^{np}t_1^i}{np}, \qquad T2 = \cfrac{\sum\nolimits_{i=1}^{np}t_2^i}{np}, \qquad \textrm{and}\qquad \cfrac{(T2-T1)}{T1}.
\end{equation}
Here, $np$ denotes the number of constrained optimization problems with similar dimensionality of a respective benchmark function set.
$T1$ and $T2$ are reported together with their relative difference $(T2-T1)/T1$.
  
  To represent a meaningful quantity of algorithm complexity $(T2-T1)/T1$, 
  the measurements $T1$ and $T2$ need to consider a sufficiently large number of function evaluations. However, such an approach can be problematic:
  Imagine a DE algorithm~\cite{TakahamaS10} (or an EDA\footnote{Please, refer to~\cite{hauschild2011introduction} for a survey about Estimation of Distribution Algorithms (EDA).  }), that initializes a rather large archive of about $200 N$ candidate solutions.
  Considering dimension $N=50$, such an algorithm would consume the whole budget of $10^4$ function evaluations in its initialization process. 
  Consequently, $T2$ cannot provide any information about the actual algorithm running time. This can be resolved by only accounting for functions evaluations consumed in the main loop of the algorithm. However, a restriction like that is not specified and would disregard preprocessing as well as initialization efforts.

 Considering the ranking of competing algorithms, the presentation style promotes the need for a well-defined algorithm ranking.
 Unfortunately, the technical report of the CEC2006 competition~\cite{CEC2006} does not provide any motivation of a suitable ranking procedure at all. 
 The presentation of the competition results is also of little help. Hence, the quality indicators used to obtain an algorithm ranking cannot be deduced.

 While defined in different ways, the ranking schemes used for the CEC2010 and CEC2017 benchmarks are fully explained.
  The CEC2010 ranking method is based on a mean value comparison of two or more algorithms on each individual constrained problem. Algorithms that yield feasibility rates of $FR=100\%$ are ordered based on their mean objective function values.
Those algorithms realizing a feasibility rate in between $0\% < FR < 100\%$ are ranked according to their feasibility rate. Finally, strategies resulting in $FR=0\%$ are ordered based on the mean constraint violations of all 25 runs.
The total rank of an algorithm is obtained by summing up its ranks on all 36 problems (including dimensions $N=10$ and $N=30$)
and the average rank $\textrm{Rank}^{CEC2010}_{avg}$ is determined by
\begin{equation}
 \textrm{Rank}^{CEC2010}_{avg} = \cfrac{\sum\limits_{i=1}^{36} \textrm{Rank}^i }{ 36}.
\end{equation}
This way the best algorithm is defined by the lowest rank value $\textrm{Rank}^{CEC2010}_{avg}$.

The CEC2017 ranking method is considering the mean objective function values as well as the median solution at the maximal allowed number of function evaluations. The first ranking of all competing algorithms is based on the mean values. 
After having completed all independent runs, for each constrained problem $i$ the algorithms are ordered with respect to their feasibility rate $FR$.
The second ordering criterion is the magnitude of mean constraint violations. 
At last, ties are resolved by considering the realized mean objective function values. Acting this way, each algorithm obtains a rank $\textrm{Rank}^i_{mean}$ on each constrained problem.
The second ranking procedure relies on the median solutions. The first ordering step is concerned with the feasibility of the median solution.
A feasible solution is better than an infeasible solution. Feasible solutions are then ordered by means of their objective function values and infeasible ones according to their mean constraint violations. On every constrained problem, each algorithm is assigned a rank $\textrm{Rank}^i_{median}$.
Having ranked all algorithms on every single constrained problem, the ranks are aggregated. That is, the total rank
value of each algorithm is calculated as 
\begin{equation}
 \textrm{Rank}^{CEC2017}_{total} = \sum\limits_{i=1}^{28} \textrm{Rank}^i_{mean} + \sum\limits_{i=1}^{28} \textrm{Rank}^i_{median}
\end{equation}
Again, the best algorithm obtains the lowest rank value $\textrm{Rank}_{total}$.

Regarding these two ranking methods, it is noticed that the CEC2017 ranking is a progression. It no longer uses a single ranking (average case quality in the broadest sense), but the consensus of average case and median case quality. This is in line with~\cite{Mersmann2015}, where the use of so-called \emph{consensus rankings} is recommended for algorithm comparison. Consensus rankings are distinguished into \emph{positional} and {optimization-based} methods. 
 \begin{table}[t] 
  \centering
  \renewcommand{\arraystretch}{1.1}{
    \begin{tabular}{rp{4.7cm}rp{4.7cm}} \hline
     \multicolumn{4}{c}{\textbf{The constrained CEC benchmarks }}  \\ \hline
 	 \multicolumn{2}{l}{\textbf{Benefits}} &  \multicolumn{2}{l}{\textbf{Areas for improvement}}  \\ 
 	+& function definitions are openly available in \texttt{C} and \texttt{Matlab}, and supported by a detailed technical report & -& the technical report needs to be more precise w.r.t. reporting duties, and box-constrained handling in particular  \\
 	+& benchmarks are frequently used in a large number of publications & -& reference data should be made available to prevent the publication of dominated results \\
 	+& great number of constrained test functions in moderate and high dimensions & -&  benchmark set lacks an allocation into problem subgroups  \\
 	+& scalable constrained functions w.r.t. the dimensionality & -& fixed number of constraints for each constrained function \\
 	+& inclusion of non-linear equality and inequality constraints & -& few problems suited for interior point strategies  \\
 	+& feasible regions of complex structure (small ratio, disconnected, etc.) & -& final ranking depends on the number of algorithms and aggregates over dimensions \\
 	+& clearly defined performance indicators & -& ranking omits algorithm speed (efficiency) \\
 	+& detailed result presentation in tabular form & -& a supporting graphical preparation of the results is omitted  \\ \hline
     \end{tabular}}
    \caption{Shorthand overview of the benefits of the CEC benchmarks for single objective constrained real-parameter optimization 
	      and related areas for improvement. More detailed statements can be found in Sec.~\ref{cec}.
	     Note that this table focuses on CEC2017 definitions that are considered representative of the predecessor versions.} 
     \label{tab05b}
 \end{table}
 
The definition of a consensus ranking is by no means unique as it is rather sensitive with respect to the choice of individual rankings and the number of considered algorithms. A desirable property of a consensus ranking would be the \emph{Independence of Irrelevant Alternatives} (IIA) criterion~\cite{Arrow1950} stating that 
changes in the number of algorithms must not affect the pair-wise preference in the consensus ranks. That is, if the consensus ranks algorithm $A1$ first and algorithm $A2$ second among five distinct algorithms, then disregarding any other algorithm should not yield a consensus rank change between $A1$ and $A2$.
However, this criterion is hardly satisfied by most intuitive consensus methods and, after all,  a ``best'' consensus ranking does usually not exist. 
Yet, a good consensus method is likely to promote insight into advantageous algorithmic ideas and might highlight poor performance, respectively. For a description and a more detailed discussion of sophisticated consensus methods, it is referred to~\cite{Mersmann2015}.

The positional consensus ranking of the CEC2017 benchmarks is created by simply adding the mean and median ranks. This can result in potentially undesirable consensus rankings\footnote{Note that the IAA criterion does not hold in this case.}. For example, consider the scenario of comparing three distinct algorithms A1, A2, A3 with mean ranking $\textrm{A1} <  \textrm{A2} < \textrm{A3}$ and median ranking $ \textrm{A3} <  \textrm{A2} < \textrm{A1}$ on a single constrained function. Consequently, all three algorithms would receive similar consensus ranks. While this may come as an exceptional case, the situation can, in fact, be observed regularly when comparing similar algorithm variants on the CEC benchmarks.
Such ties can be resolved by including a third ranking approach into the consensus method. 
As the CEC2017 rankings do not address any measure of algorithm efficiency, a third ranking might take into account the algorithm speed in terms of function evaluations. A possible step in this direction could be a ranking that is based on the observed mean and median ranks realized after having consumed $10\%$, and $50\%$, of the evaluation budget.

Moreover, the CEC ranking approaches aggregate algorithm rankings over multiple dimensions. This way, algorithms which are especially well performing in lower dimensions are potentially overrated and the overall ranking might be prejudiced. Further, algorithms that are particularly well performing in larger dimensions cannot be clearly identified. Aggregation over dimension should be avoided because the problem dimension is a parameter known in advance that can and should be used for algorithm design decisions~\cite{hansen2016perf}.

To conclude this review of the constrained CEC benchmarks, some of the mentioned aspects could be incorporated in the advancing CEC competitions on constrained real-parameter optimization. In doing so, algorithm developers would benefit from the introduction of well-designed problem subgroups that support the identification of particularly difficult problem features. Further, competing algorithms should be ranked for individual dimensions in order to obtain an intuition of the scalability of an algorithm. A competition winner might then be assigned by weighting these ranks. 

 Table~\ref{tab05b} recaps the vital benefits of the CEC benchmarks as well as some room for improvement which has been mentioned in more detail within this and the previous subsections. However, not all of the mentioned improvements can automatically be considered a shortcoming of the CEC benchmarks. The benchmarks might rather be based on design decisions with different emphasis.

\section{The COCO framework}
\label{coco}

   The Comparing Continuous Optimizer (COCO) suite~\cite{hansen2016coco} provides a platform to 
benchmark and compare continuous optimizers for numerical (non-linear) optimization.
Only recently, the development of a COCO branch for constrained optimization problems started. 
The related code is available on the project website\footnote{\url{https://github.com/numbbo/coco} The corresponding documentation is provided in~\cite{sampaio2016coco}  under \url{docs/bbob-constrained/functions/build} after building it according to the instructions.} within the \texttt{development} branch.
For convenience, the COCO benchmark test suite for constrained functions is synonymously referred to as COCO BBOB-constrained, or simply COCO, respectively.

While the COCO BBOB-constrained testbed is not yet operational, being short before completion, the corresponding benchmarking principles and the associated test problem structure are not expected to substantially change anymore. As the COCO framework represents the currently most elaborated benchmarking environment for EAs, not mentioning the constrained COCO principles would render the present review incomplete.

However, caution is advised with respect to small changes in individual test function aspects, e.g. the distances of the constrained optimal solution from the unconstrained optimal solution\footnote{Note, the unconstrained optimal solution of a constrained function~\eqref{cop} is associated with the optimum of the related objective function, i.e. disregarding all constraints.} or regarding the post-processing practice.

The rest of this section is concerned with pointing out the COCO BBOB-constrained benchmarking conventions, the related test problem definitions, the evaluation criteria as well as the presentation style.

\subsection{Benchmarking principles}
\label{cocofund}
The COCO BBOB-constrained suite is distinctly built on the unconstrained COCO framework. 
The COCO platform assists algorithm engineers in setting up proper experiments for algorithm comparison.
It provides simple interfaces to multiple programming languages (C/C++, Python, MATLAB/Octave, and Java) which makes the benchmarks easily accessible.
Users are not involved in the evaluation of constrained functions or the logging process of algorithm results.
A corresponding post-processing module facilitates the illustration and the meaningful interpretation of the collected algorithm data.
In this respect, COCO reduces the benchmarking effort for algorithm developers with respect to implementation time.

The benchmark functions are considered to represent black-box functions for the tested algorithms. Still, the objective functions are explicitly stated in mathematical form in the documentation. This allows for a deeper understanding of the individual problem difficulties and thus of an algorithm's (in)capabilities.
In a first step, the COCO BBOB-constrained test bed confines itself to eight well-known objective functions from the context of the unconstrained COCO suite. These objective functions are provided with varying number of (almost) linear inequality constraints.
However, the actual test instances are randomly generated for each algorithm runs, see Sec~\ref{cocoexp}.
A very comprehensive explanation of the COCO framework and the associated constrained problems can be found on the COCO documentation website\footnote{\url{https://numbbo.github.io/coco-doc/}}.

The COCO guideline for counting function evaluations in the constrained setting involves distinguishing objective function evaluations and constraint evaluations.
Still, one constraint evaluation is identified with the evaluation of all individual constraint functions at a time. 
Accordingly, a specified budget of function evaluations needs to be split.

On the one hand, the formal constrained function definitions are not specifying any box-constraints, refer to~\eqref{copcoco}. In this regard, guidelines for the treatment of box-constraints are not needed. Yet, the BBOB-constrained suite provides the user with the subroutines \texttt{cocoProblemGetSmallestValuesOfInterest}, and \texttt{cocoProblemGetLargestValuesOfInterest}, to determine the lower bound $\bm{\check{y}}$ and the upper bound $\bm{\hat{y}}$ for each constrained problem. While the optimal solution is located inside the box $\mathcal{S}$ according to Eq.~\eqref{boxset}, evaluations of candidate solutions outside the box are not interdicted. 

Whether the box-constraints need to be enforced in every step or not is of course a design question. Anyway, the benchmark designers need to provide plain instructions with respect to treatment of box-constraints during the search process. The use of the box-constraint handling may be beneficial on some constrained problems. 
Therefore, such instructions are necessary to obtain comparable algorithm results. 
Moreover, algorithm developers need to be urged to report the specific box-constrained handling techniques used.

\subsection{Experimental design}
\label{cocoexp}

The standard BBOB-constrained optimization problem reads
\begin{equation}
      	\begin{aligned}
	\min \:\: & f(\bm{y})  & &\\
	  s.t. \:\: & g_i(\bm{y}) \leq 0,  & i=1,\dots,l,\: &\\
	      & \bm{y} \in \mathds{R}^N. & &
	\end{aligned}
	\label{copcoco}
      \end{equation}
A summary of the associated problem features is provided in Table~\ref{tab06}.      
The considered constrained functions are separated into eight subgroups associated with the selected objective functions. These objective functions are
    \begin{itemize}
	\itemsep0pt 
	\item the Sphere function, 
	\item the Ellipsoid function, 
	\item the Linear slope function,	    
	\item the rotated Ellipsoid function,
	\item the rotated Discuss function, 
	\item the rotated Bent Cigar,  
	\item the rotated Different Powers, and  
	\item the rotated Rastrigin function.  
  \end{itemize}
By systematically equipping each objective function with 6 different numbers of inequality constraint functions, namely $1$, $2$, $6$, $6 + \lfloor \frac{N}{2} \rfloor$, $6 + N$, and $6 + 3N$ constraints, the BBOB-constrained benchmark problems are built.
The number of the constraints depends on the considered search space dimension $N$.
Note that the BBOB-constrained suite renounces the incorporation of equality constraints.
\begin{table}[t] 
  \centering
  \renewcommand{\arraystretch}{1.1}
    \begin{tabular}{p{6cm}p{5cm}}
 	Benchmark name & COCO BBOB-constrained \\ \hline \hline
 	Search space dimensions  & $N \in \{2, 3, 5,10,20,40\}$   \\[0.5ex]  \hline
 	Number of constrained functions   & $288$ (incl. varying dimensions) \\[0.5ex]
 	Number of distinct obj. functions & $ 8$   \\[0.5ex]
 	Minimal number of constraints & $1$ \\[0.5ex]
 	Maximal number of constraints & $126$\\[0.5ex]\hline
	Scalable problems    &  yes \\[0.5ex]
	Budget of function evaluations  & user-dependent  \\[0.5ex]
 	Number of fully separable problems (objective and constraints) & $ 144 $\\[0.5ex] \hline
	Avg. size of $\rho = \mathcal{M}/\mathcal{S}$ & $21.1\%$  \\[0.5ex]
	Number of problems with $\rho > 10^{-3}$ & $200$ \\
     \end{tabular}
    \caption{Characteristic features of the COCO BBOB-constrained benchmark suite.} 
    \label{tab06}
 \end{table}
 
For now, the COCO BBOB-constrained testbed concentrates on \emph{almost} linear inequality constraints.
To this end, the linear structure of the feasible region is distorted by application of bijective non-linear transformations on a number of constrained functions.
The subsequent application of a randomly generated translation of the whole constrained problem prevents the optimal solution from being the zero vector, i.e. $\bm{y}^* \neq \bm{0}$. 
The problems are further created in a way that maintains a known optimal solution of the constrained function.
This optimal solution is always located on the boundary of the feasible region.  
However, considering the black-box setting the optimal solution is not accessible by a user, nor by the algorithm.
It is used for evaluation of algorithm performance.

The procedure to create a constrained function consists of five steps\footnote{The construction of the constrained Rastrigin function group is slightly different. It is referred to~\cite{atamna2017,sampaio2016coco} for the detailed definition.}:
 \begin{itemize}
		  \item[I.] Select a pseudo-convex objective function $f(\bm{y})$ and a corresponding number $l \in \{1,2,6,6+\lfloor \frac{N}{2} \rfloor, 6+N, 6+3N\}$ of constraints $g_i(\bm{y}),\: i=1,\dots,l$.
		  \item[II.] Define the first linear constraint $g_1(\bm{y}) \coloneqq -\nabla f(\bm{0})^\top \bm{y}$. 
		  \item[III.] Construct the remaining linear constraints $i=2,\dots,l$ by sampling their gradients from a multivariate normal 
			      distribution and incrementally demanding that the origin remains a Karush-Kuhn-Tucker (KKT) point of the problem~\cite{boyd2004convex}. 
		  \item[IV.] If applicable, apply non-linear transformations to the constrained function.
		  \item[V.] Randomly sample a translation vector to change the location of the optimal solution.
		\end{itemize}

According to the COCO BBOB-constrained documentation~\cite{sampaio2016coco}, the domain of almost linear constrained functions represents the most interesting starting configuration for benchmarking.\footnote{While this can be disputed, it is likely the most simple and logical step for gradually extending the COCO BBOB framework to the constrained problem domain.} Such constraint functions are composed of small variations of linear constraints which are considered to represent most simple restrictions to an unconstrained optimization problem.
Algorithms suitable for constrained optimization, in general, should first be able to solve such (almost) linearly constrained functions.
The inequality constraints are considered to be relaxable, i.e. candidate solutions outside of the feasible region can be evaluated and may contribute to the search process of an algorithm.

The transformations are essentially applied to ensure constrained functions that are reasonably difficult to solve, i.e. potential regularities that might favor the exploitation abilities of certain algorithms are excluded. The transformations are designed in such a way that the automatic generation of similarly hard test problem instances is realized.
Problem instances share the objective function, the number of inequality constraints, as well as the search space dimension. 
By randomly defining and distorting the linear constraints, the size of the feasible region may vary. 
The extent to which the complexity of two instances with differently sized feasible regions is maintained remains unanswered.

The constrained problems~\eqref{copcoco} are scalable with respect to search space dimension $N$ and number of constrained functions $l$.
Taking into account dimensionality, objective function and the number of constraints, the BBOB-constrained testbed consists of $288$ distinct constrained functions.
By composing problem subgroups by means of objective functions, as well as dimensionality, supports the identification of algorithmic strengths and weaknesses for specific problem characteristics.
The constrained COCO framework considers only inequality constraints $g_i(\bm{y})\leq 0$. Consequently, a candidate solution is regarded feasible solution if all inequality constraints are satisfied, i.e. $g_i(\bm{y})\leq 0 \quad \forall i=1,\dots,l$. By construction, the feasible sets of the benchmark suite is non-empty and connected.
For initialization purposes a feasible candidate solution is provided by the COCO subroutine \texttt{cocoProblemGetInitialSolution}. It may serve as a starting point for the search process. 
This represents a beneficial feature for benchmarking algorithms that search exclusively inside the feasible region $\mathcal{M}$, see~\eqref{feasSet}.
As already mentioned in Section~\ref{cocofund}, the box constraints of each problem are accessible. 
Hence, they may also be used to initialize a starting population inside the box $\mathcal{S}$. 

When estimating the size of the feasible region relative to the box defined by the lower and upper bounds, the associated $\rho$ values indicate the dependence of the dimension $N$. 
Yet, the aggregated $\rho$ value presented in Table~\ref{tab06} only has limited significance. On the one hand, it was generated according to~\cite{Koziel1999} by considering only a single instance of each constrained function. As the randomly generated boundary of the feasible region may vary among constrained problem instances, the $\rho$ value is supposed to exhibit fluctuations of some degree. On the other hand, the $\rho$ was averaged over all possible problem dimensions and thus only represents a rough sketch.
However, compared to the CEC benchmarks in Section~\ref{cec}, the average feasible region of a BBOB-constrained function can be considered larger.

The benchmark suite does not determine a fixed budget of function evaluations. The specification of appropriate termination conditions for an individual algorithm is left to the user~\cite{HansenTMAB16}. In this context, the COCO built-in function \texttt{cocoProblemFinalTargetHit} delivers an indicator of the realized algorithm precision. It returns \texttt{true} after the algorithm has approached the optimal objective function value with accuracy $10^{-8}$ and can be utilized to terminate the algorithm run. Accordingly, the value of $10^{-8}$ represents the final target precision that is used to specify a successful algorithm run, see Section~\ref{cocorep}.

By default, each algorithm is executed on $15$ randomly generated instances of each constrained function.
The corresponding results are interpreted as 15 independent repetitions on the same constrained problem. 
Acting this way prevents unintentional exploitation of potentially biasing function features~\cite{hansen2016coco}.

Remember that the optimal solution is by construction located on the boundary of the feasible region. This property might potentially prejudice search algorithms to  largely operate outside of the feasible region of the search space. Depending on the fitness environment, this allows for faster progress until the algorithm reaches a certain neighborhood of the optimal solution. 

\subsection{Reporting}
\label{cocorep}

The COCO framework comes with a post-processing module for automated data preparation and visualization in terms of \texttt{html} or \texttt{LaTeX} templates.
The user-independent standardization of the data processing reduces the susceptibility to errors and supports the comparability of algorithm performance.

The COCO BBOB-constrained suite takes into account a single performance measure: the algorithm runtime.\footnote{By concentrating on runtime, the BBOB-constrained benchmarks may refrain from defining an order relation for candidate solutions.}
Runtime is defined in terms of the number of function evaluations\footnote{Keep in mind, that the number of function evaluations comprises the sum of all objective function evaluations \emph{and} the number of constraint evaluations.} consumed on a specific constrained problem until a predefined target is reached.
In total, $51$ targets uniformly distributed on the log-scale are specified in the range $[10^2,10^{-8}]$ individually for each constrained function. 
More specifically, for each of these target values, the runtime of an algorithm is identified with the number of objective function and constraint evaluations consumed until a target was reached for the first time.
In the case that not all targets are reached by an algorithm, the COCO framework makes use of a bootstrapping method~\cite{efron1994introduction}.
This method permits to compare algorithms with different success rates. A detailed description is available in~\cite{hansen2016perf}.

Whether a target was reached after evaluation of a candidate solution is automatically checked by the COCO suite. 
To this end, a trigger value is compared with the next unmatched target.
The corresponding number of function evaluations as well as the trigger value are logged. 
For now, the trigger value is identified with the objective function value of a feasible candidate solution. Infeasible candidate solutions, or their constraint violations, are not considered in the definition of the currently used triggers. 
The objective function value of the initially provided solution \texttt{cocoProblemGetInitialSolution} is considered as initial trigger value.
The initial trigger value does usually not satisfy any of the targets. It is updated as soon as the benchmarked algorithm is able to find a feasible candidate solution with improved objective function value.
\begin{figure}
\centering
 (a)\includegraphics[trim=10 0 30 0,clip,width=0.75\textwidth,height=0.55\textwidth]{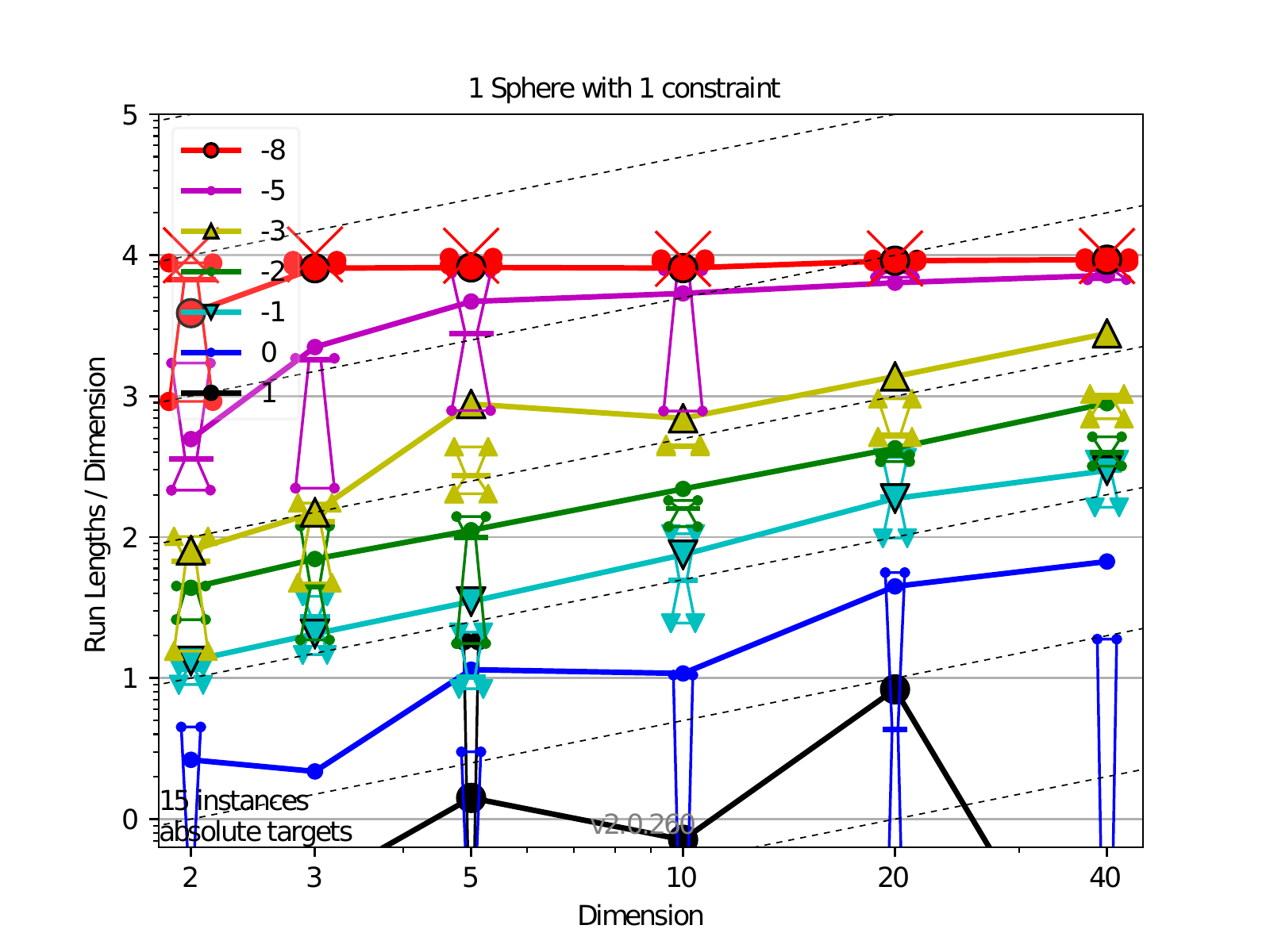} \\(b)\includegraphics[trim=10 0 30 0,clip,width=0.75\textwidth,height=0.55\textwidth]{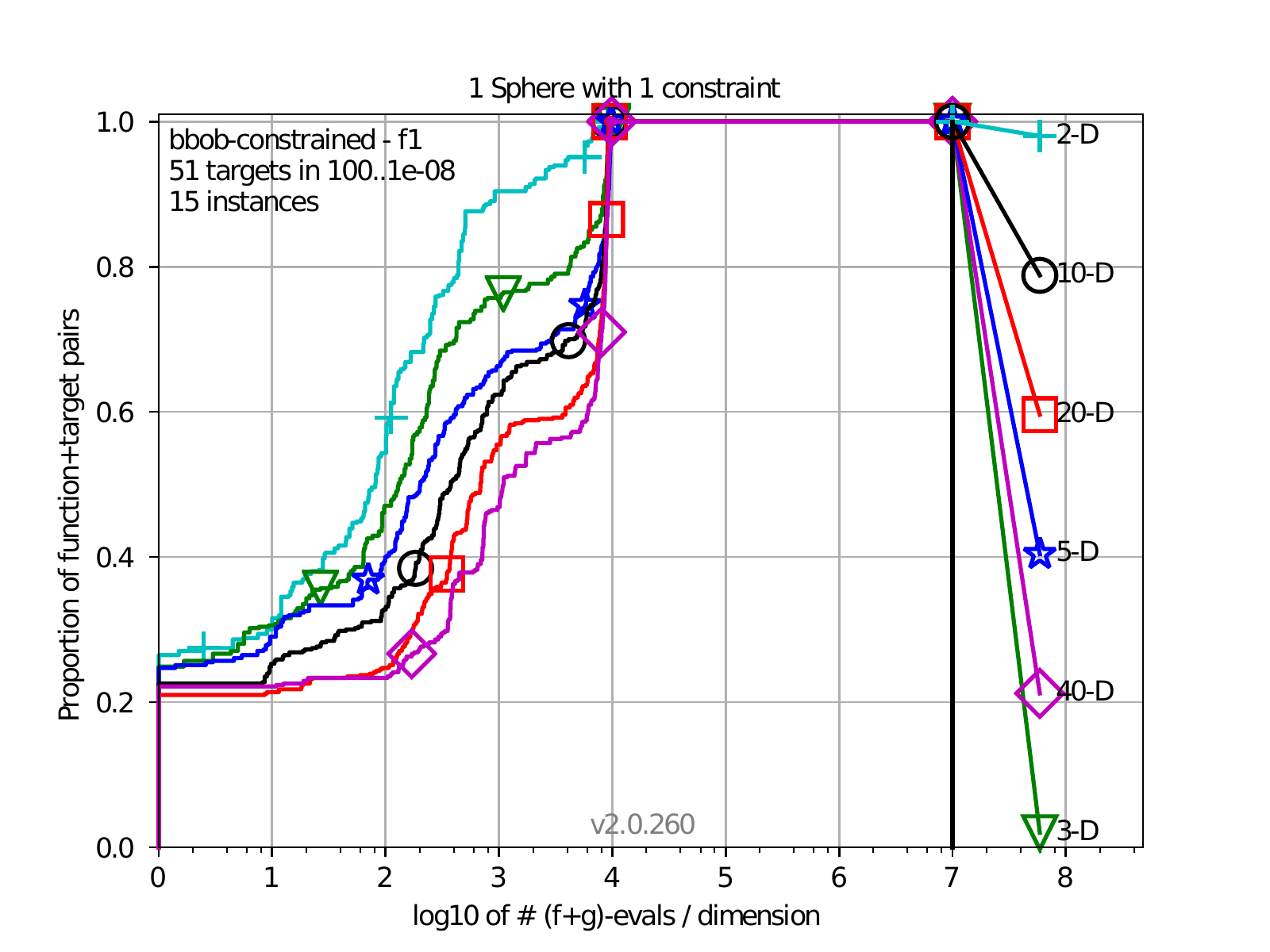}
 \caption{An excerpt of the presentation style of benchmarking results obtained on the first problem of the BBOB-constrained suite. 
	  (a): Scaling of runtime with dimension to reach certain target values on the BBOB-constrained benchmarks. Lines: average runtime (aRT); 
	    All values are divided by dimension and plotted as $\log_{10}$ values versus dimension.
	    Shown is the aRT for fixed values of $\Delta f = 10^k$ with $k$ given in the legend. \: (b): Bootstrapped empirical cumulative distribution of the number of objective function and constraint evaluations divided by dimension for 51 targets with target precision in $10^{[-8\dots2]}$ for dimensions $2,3,5,10,20$, and $40$. The horizontal axis shows the $\log_{10}$ of the sum of objective function and constraint evaluations. The vertical axis shows the proportion of target objective function values reached with the given number of evaluations. 
	    \\
	    Note, this caption has been adapted from the COCO BBOB post-processing \LaTeX template.
	    }
 \label{ECDFplot}
\end{figure}

Making use of this runtime definition results in a performance measure that is essentially independent of the computational platform and the programming language used.
Further, the algorithm results can easily be condensed and presented in multiple ways, e.g. by measuring the average runtime (aRT) of an algorithm~\cite{hansen2016perf}, by use of data profiles or empirical cumulative distribution function (ECDF) plots~\cite{MoreWild2009}, or runtime tables for specific target values.
An illustration of an aRT plot is displayed in Figure~\ref{ECDFplot}(a). It provides an estimate of the expected runtime. The aRT is computed by summing up all evaluations in unsuccessful algorithm runs as well as the number of evaluations consumed in the successful algorithm runs, both divided by the number of successful runs. 
The ECDF plot provided in Fig.~\ref{ECDFplot}(b) displays the proportion of successfully reached targets on function \textbf{f01} plotted against the number of function evaluations.It is usually independent of any reference algorithms and thus unconditionally comparable across different publications.
This supports drawing meaningful conclusions with respect to algorithm performance on the whole benchmark set, or on the individual problem subgroups, respectively.
Note, that algorithm results are not aggregated over dimensions in order to disclose the impact of the problem dimensionality on the algorithm performance.

Algorithms can be directly compared by illustrating their ECDFs per function evaluations in log-scales. This way, the area above and in between the graphs 
becomes a meaningful conception. An exemplary ECDF is illustrated in Figure~\ref{ECDFplot}. It can be interpreted in two ways: By considering the number of function evaluations on the $x$-axis as independent variable, the $y$-axis represents the ratio of targets reached for any budget $x$. On the other hand, associating the $y$-axis with the independent variable, the $x$-values present the maximal runtime observed to reach any fraction $y$ of the predefined target values.

Consequently, better performing algorithms realize smaller areas above a curve. Further, the difference between those areas can be interpreted as a measure of the performance advantage of one algorithm over another. 
With the caveat of loosing the connection to a single constrained problem, the ECDF plots allow for aggregation over multiple constrained problems~\cite{hansen2016perf}. 
That is, the presentation of algorithm performance on problem subgroups is straight forward. 
Hence, in contrast to extensive and hardly interpretable tables, the ECDFs provide a relevant notion of algorithm suitability for single constrained functions, and subgroups of constrained problems, respectively.\footnote{The ECDF aggregation over different dimensions is omitted to prevent loss of information related to the impact of the search space dimension on the algorithm performance.}

Only considering feasible candidate solutions in the trigger/target definition may inflate the relevance of late phases in the search process.
Depending on the constrained problem, algorithms that sample an initial population within the box-constraints might consume a considerable number of function evaluations until they reach the feasible region.  The number of function evaluations needed to hit a first target provides a notion of the runtime needed to find a first feasible solution. Accordingly, the area to the left of a ECDF curve can still be identified with the runtime of the respective algorithm.
However, the resulting ECDF plots will thus likely display a steeply ascending curve that is shifted to the right boundary (determined by the limit of function evaluations). This complicates the comparison of multiple algorithms because the relevant information might be largely accumulated in one spot.
Also from a practical point of view, the late search phase may have minor impact on the assessment of an algorithm if the main focus is on finding a feasible solution of reasonable precision.  

Other trigger definitions are conceivable, i.e. the trigger may be defined by the sum of the objective function value and the constraint violation of a candidate solution. This way of proceeding takes into account infeasible steps, but it would introduce the issue of unwanted cancellation effects.
Another idea to give an impression of the algorithm performance within the infeasible region is the definition of separate targets for the constraint violation. These targets would need to be displayed in a second plot that addresses the runtime during the search in the infeasible region of the search space.
 \begin{table}[t] 
  \centering
  \renewcommand{\arraystretch}{1.1}{
    \begin{tabular}{rp{4.7cm}rp{4.7cm}} \hline
 	      \multicolumn{4}{c}{\textbf{The COCO BBOB-constrained benchmarks }}  \\ \hline
 	 \multicolumn{2}{l}{\textbf{Benefits}} &  \multicolumn{2}{l}{\textbf{Areas for improvement}}  \\  
 	+& function definitions openly available in \texttt{C}, \texttt{Java}, \texttt{Python} and \texttt{Matlab} & -& technical report needs to be more precise w.r.t. reporting duties  \\
 	+& detailed motivation and construction of the experiments & -& no statement on the treatment of box-constraints \\
 	+& scalable constrained functions w.r.t. the dimensionality and number of constraints & -&  limitation to 8 objective function definitions and (almost) linear inequality constraints \\
 	+& initial feasible solution available (suited for interior point strategies) & -& only constrained functions with connected feasible regions included \\
 	+& clearly defined performance measure (objective function targets) & -&  optimal solution always located on the boundary of the feasible region \\
 	+& algorithm efficiency measured based on consumed function evaluations per target & -& only feasible targets defined (performance within infeasible region ignored)\\
 	+& standardized post-processing and data visualization by use of ECDF plots & -& supportive tabular presentations are omitted  \\ \hline
     \end{tabular}}
    \caption{Summary of the features of the COCO BBOB-constrained benchmarking definitions mentioned in Sec.~\ref{coco}.} 
 
    \label{tab06b}
 \end{table}

The COCO experiments include the approximate measurement of the algorithm time complexity~\cite{HansenTMAB16}. To this end, it is recommended to monitor either the wall-clock or the CPU time while running the algorithm on the benchmark suite. The time normalized by the number of function evaluations is demanded to be reported for each dimension. Additionally, information on the experimental setup, the programming language, the chosen compiler and the system architecture are required. Yet, the instructions do not fully exclude diverse interpretations and may thus impede the comparability and reproducibility of the results.

As the development of the BBOB-constrained benchmark suite is still ongoing, the definitive presentation style of the algorithm results cannot be provided at this point. The presentation of additional information on the ratio of the feasible region relative to the box $\mathcal{S}=[-5,5]^N$ is conceivable.
Further, the ultimate choice of the trigger value for deciding whether a predefined target was reached is still being discussed.\footnote{For the ongoing discussion on BBOB-constrained features, it is referred to~\url{https://github.com/numbbo/coco/issues}.} 
 
 To wrap up the deliberations of Sec.~\ref{coco}, Table~\ref{tab06b} provides a shorthand overview of the benefits, as well as potential areas for improvement, of the BBOB-constrained testbed. It has to be noted that the mentioned improvements can not necessarily be regarded as a shortcoming of the respective benchmarking environment, as they might represent reasonable design decisions in the process of advancing towards a well-elaborated benchmarking environment.

  \section{Conclusion}
 
    \label{discussion}
       
    The present review intends to collect principles for comparing constrained test environments for Evolutionary Algorithms. 
    To this end, it takes into account recommendations on the basic principles, the experimental design, and the presentation of algorithm results.
    Based on the gathered criteria, the most prominent constrained benchmarking environments for EAs are reviewed. 
    Significant differences with respect to the basic assumptions and the experimental approaches became evident.
    The survey of the current constrained benchmarking sets suitable for randomized search algorithms supports the algorithms developers with 
    information about the strength of the available frameworks.

    Both considered benchmark suites focus on different constrained problem domains. They differ in terms of counting function evaluations, defining termination criteria as well as performance evaluations comparison. The COCO BBOB-constrained benchmark is very much based on the unconstrained COCO framework.
    By including exclusively almost linear inequality constraints, it represents a first systematic attempt towards general constrained problems. 
    The BBOB-constrained test function definitions are rather tangible. This is due to the composition of well-known unconstrained optimization problems and connected feasible sets. By construction, the BBOB-constrained benchmarks (internally) maintain an optimal solution for measuring algorithm performance.
    In comparison, the structure of the constrained CEC test problems is somehow harder to perceive.
    Being also based on proven unconstrained objective functions, the structure of the corresponding feasible sets is comparably complex.
    A reason is varying numbers of usually non-linear equality and inequality constraints that potentially define disjoint feasible regions in the search space. 
    Further, the most recent constrained function definitions do not provide information about optimal solutions.
     
    These distinct benchmarking approaches directly induce different ways of presentation. On the one hand, the COCO framework measures runtime in terms of function evaluations per predefined target and visualizes algorithmic performance in terms of ECDF graphs. Algorithm performance can thus be rather easily aggregated over similar problems and compared to different algorithms. On the other hand, the CEC benchmarks compute a number of best quality, median quality, or mean quality indicators and illustrate the algorithm performances by use of tables. The assessment of algorithmic ideas is rather cumbersome. Further, the comparison of algorithm results thus relies on ranking schemes that may come with a sense of arbitrariness.
        
    Both, the CEC competitions for constraint real-parameter optimization, and the COCO BBOB-constrained framework do only consider relaxable equality and inequality constraints. 
    That is, the algorithms are allowed to move in the whole unconstrained search space. Each candidate solution, either feasible or infeasible, may be evaluated and used within the variation or selection steps of the strategy. 
    It should be reminded that the possibility to use infeasible solutions during the search may significantly reduce the problem complexity.
    An algorithm might completely operate outside the feasible region until it finds the optimal solution. 
    Taking into account the size of the feasible regions, and looking at the problem definitions of the CEC 2006, 2010, and 2017 benchmarks, 
    it should be made clear that many problems have disjoint feasible regions. 
    Hence, enforcing the feasibility of candidate solutions prior to their evaluation appears useless on the CEC benchmarks.
    However, the prior demand for feasibility is often required in real-world problems, e.g. when considering simulations which require feasible inputs. In this regard, the CEC benchmarks do not represent a suitable test function class.
    Similar concerns can be raised for the BBOB-constrained benchmark suite. However, its feasible set is always non-empty and connected. Providing an initially feasible solution, the BBOB-constrained framework could potentially take into account unrelaxable constraints.
 
    Furthermore, both benchmark sets omit to demand a specific box-constraint treatment. Yet, they refrain from mentioning the need of the precise reporting of such approaches. Considering the source codes of the most successful strategies reported in CEC competitions, all algorithms were assuming situation (S1).
    Even if not explicitly specified in the benchmark definitions, today the enforcement of the bound constraints seems to be 'common sense' within the Evolutionary Computation community. However, the mechanisms to treat box constraint violations may vary and are usually not well reported.
    As pointed out in Sec.~\ref{boxconstr}, plain instruction with respect to the treatment of box-constraints can prevent inconsistencies~\cite{LiaoMMS2014}.
    
    Considering the CEC benchmark environments, the problem definitions were subject to considerable changes in recent years. 
    The introduction of scalable constrained functions was accompanied by a reduction of the average number of constraints per problem (from $7$ to about $2$).
    While the CEC2006 benchmarks were (partly) inspired by real-world applications, the comparably small fixed number of $2$ constraints appears underrepresented when taking into account the structure of real-world problems.
    Further, parameter space transformations were introduced in order to remove potential problem biases in direction of the coordinate axes. 
    Still, a small number of fully separable constrained functions remained in the CEC2017 benchmark set. 
   
    The constrained CEC benchmarks provide the currently most elaborated benchmarking environment for EA.
    They mainly present non-linearly constrained problems with a fixed number of not necessarily linear inequality and equality constraints. Also due to unconnected feasible sets, the CEC constrained test functions are considered to represent hard challenges in some cases.
    Yet, further improvements are still conceivable (refer to Table~\ref{tab05b}).    
    On that note, a comprehensive documentation that motivates the advancement of the constrained CEC benchmark environments is missing.
    Future CEC benchmarking competitions also might consider providing a repository of baseline algorithm results in order to assess the competitiveness of algorithmic ideas and to highlight actual advancements in this field of research.
    
    Looking at the recent CEC2017 benchmarking functions, the distinct problem features introduce a rather high level of problem complexity. The benchmarks are suited to demonstrate the use of algorithmic ideas. According to its intention in the context of the CEC competition, the CEC benchmarking environment is well designed to assess algorithmic performance and compare algorithms over a broad range of different constrained test problems. 
    Yet, the lack of problem subgroups complicates the identification of correlations between successful algorithmic working principles and specific problem features. In this respect, the benchmark set does only weakly support the iterative development process of specialized algorithms for particular constrained problems.      
    
    Regardless of minor software bugs and unfinished post-processing methodology, the
    COCO BBOB-constrained suite could have the potential to become another standard constrained benchmarking platform. It is equipped with a detailed documentation of its benchmarking principles as well as an elaborated post-processing strategy. The experimental design of the COCO BBOB-constrained benchmarks advances well-known unconstrained test functions to the constrained problem domain.  To this end, each objective function is accompanied with a scalable number of almost linear inequality constraints.             
    The potential of BBOB-constrained is supported by the COCO framework representing a widely accepted benchmarking suite for the unconstrained case.
    The recent collection of well-structured test functions must be regarded as a reasonable first step towards an elaborate constrained benchmarking testbed. Its structure supports the development of EA variants suitable for selected problem groups. Due to its scalability, the impact of the number of constraints on the performance of algorithmic ideas can be assessed.
    
    However, comprising only a somehow limited number of distinct constrained problem types, algorithms that perform well on the COCO BBOB-constrained suite are not guaranteed to be successful on other constrained problems. In this respect, BBOB-constrained needs to proceed towards more complex constraint definitions of different types, e.g. non-linear inequality and equality constraints. Hence, the COCO constrained benchmarks should be advanced once the first version of the testbed is released (also refer to Table~\ref{tab06b}). 
    
    The COCO BBOB-constrained problem definitions might further permit (limited) user customizations. For instance, it should be possible to optionally move the optimal solution from the boundary into the feasible region. This could potentially increase the problem complexity for some constrained functions. Another option that could be thought of is manually turning off the non-linear perturbations for all constrained functions. This would result in linearly constrained problems (with non-linear objective functions) and might be useful for examining specific algorithmic ideas suited for constrained problems that lack appropriate benchmarking environments. 
    
    Taking into account the vast number of constrained problem characteristics, the current benchmarking environments under review do only cover a small ratio of the constrained problem domain. 
    Both benchmarking environments might be extended with additional constrained test functions.
    However, a drawback of extending the number of scalable problem subgroups within a benchmarking testbed is the increasing computational effort.
    Moreover, the tangibility of the chosen presentation style may be significantly reduced when considering too many different problem representations.
    This might be circumvented by embracing multiple coexisting benchmarking environments that are specialized in well-defined constrained problem classes.
    
    Aiming at the establishment of profound benchmarks for real-valued constrained optimization, the two approaches should not be regarded as opposing but rather as complementing benchmarking suites. Both environments are based on reasonable design decisions that might need an upgrade but cannot be fully negated. 
    Ultimately, an algorithm developer has to choose the benchmarking environment that best suits the application area of interest. 
    That is, highly specialized algorithms may only be practical on constrained problem subgroups which are only available in one benchmarking environment.
    Contrary, algorithms designed to be successful on a preferably broad range of problems should be assessed and compared on both (all) available constrained benchmarking environments. Accordingly, the current benchmarks for constrained single objective real-parameter optimization do support each other.

    In the end, benchmarking environments have to demand diligent scientific investigations. In particular, algorithm developers must be urged to maintain reproducible and comparable algorithm results. Collecting principles for elaborate constrained benchmarking, Section~\ref{sec3} can be regarded as a guideline for design conventions and reporting obligations.
    
    Advancing the CEC benchmark definitions, and finishing the COCO BBOB-constrained benchmark suite, are anticipated tasks for future research. 
    Further the design of additional EA benchmarking tools for different constrained problem sub-domains needs to be challenged. 
    A possible step in this direction might be the consideration of linear constrained optimization problems suited for EA. 
    In this regard, the Klee-Minty problem is able to serve for demonstrating and examining the capabilities of EA in the context of linear optimization. It is based on the Klee-Minty polytope~\cite{klee1970good}, a unit hypercube of variable dimension with perturbed vertices, which represents the feasible region of the linear problem. The linear objective function is constructed in such a way that the Simplex algorithm yields an exponential worst-case running time. 
    Considering the number of sophisticated deterministic approaches available, taking into account linear optimization problems for EA benchmarking may appear questionable in the first place. However, many purpose-built algorithms for linear optimization~\cite{IPM1,IPM2} show poor performance in this environment.
    The Klee-Minty problem was already used to compare a specially designed CMSA-ES variant for linear optimization with open source interior point LP solvers in~\cite{Spettel2018}. 
    	
In case that this review fosters the impression of an unbalanced criticism, this conjecture is probably due to the fact that the constrained CEC benchmarks have existed for many years providing a multitude of benchmarking papers and working points, respectively. In contrast, there are hardly any algorithm comparisons that were carried out on the basis of the BBOB-constrained environment. In this respect, the COCO BBOB-constrained framework will have to prove itself in practice.

    \section*{Acknowledgement}
    This work was supported by the Austrian Science Fund FWF under grant P29651-N32.


\end{document}